\documentclass{applemlr}

\usepackage{amsmath}
\usepackage{enumerate}
\usepackage{algorithm}
\usepackage{algpseudocode}
\usepackage{amsfonts}
\usepackage{amsthm}
\usepackage{cleveref}
\usepackage{diagbox}
\usepackage{colortbl}
\usepackage{amssymb}
\usepackage{xspace}
\usepackage{wrapfig}
\usepackage{adjustbox}
\usepackage{tabularx}
\usepackage{booktabs}
\usepackage{mathtools}
\usepackage{tikz}
\usepackage{enumitem}
\usepackage{silence}
\usepackage{dsfont}
\usepackage[table]{xcolor}
\usepackage[dvipsnames]{xcolor}
\usepackage{multirow}
\usepackage{makecell}
\usepackage{xfakebold}

\usepackage{amsmath,amsfonts,bm}

\def\eqref#1{equation~\ref{#1}}

\def\1{\bm{1}}

\DeclareMathAlphabet{\mathsfit}{\encodingdefault}{\sfdefault}{m}{sl}
\SetMathAlphabet{\mathsfit}{bold}{\encodingdefault}{\sfdefault}{bx}{n}

\definecolor{textgray}{HTML}{6E6E73}
\usetikzlibrary{positioning, calc}
\usetikzlibrary{decorations.pathmorphing}

\makeatletter
\patchcmd{\wrong@fontshape}{\@gobbletwo}{}{}{}
\makeatother
\numberwithin{equation}{section}
\makeatletter
\AtBeginDocument{
  \urlstyle{sf}
  
}
\makeatother

\definecolor{light}{RGB}{125, 125, 125}
\crefname{tcb@cnt@pbox}{code}{code}
\Crefname{tcb@cnt@pbox}{Code}{Code}
\crefname{assumption}{assumption}{assumption}
\Crefname{assumption}{Assumption}{Assumptions}

\newtcolorbox[auto counter]{pbox}[2][]{
  colback=white,
  title=Code~\thetcbcounter: #2,
  #1,fonttitle=\sffamily,
  fontupper=\sffamily,
  arc=2pt,
  colframe=bgcolor,
  coltitle=fgcolor,
  colbacktitle=bgcolor,
  toptitle=0.25cm,
  bottomtitle=0.125cm
}

\makeatletter
\newcommand\applefootnote[1]{%
  \begingroup
  \renewcommand\thefootnote{}%
  \renewcommand\@makefntext[1]{\noindent##1}%
  \footnote{#1}%
  \addtocounter{footnote}{-1}%
  \endgroup
}
\makeatother

\definecolor{cverbbg}{gray}{0.90}

\usepackage{xspace}
\usepackage{xcolor}
\usepackage{multirow}
\usepackage{booktabs}
\usepackage{wrapfig}
\usepackage{natbib}

\usepackage{siunitx}
\usepackage{tocloft}
\usepackage{etoc}
\usepackage{titletoc}
\newcommand\DoToC{%
  \startcontents
  \printcontents{}{1}{\noindent \textbf{\Large{Table of Contents in Appendix}}\vskip3pt\vskip5pt}
  \vskip3pt\vskip5pt
}

\usepackage[most]{tcolorbox}
\definecolor{mylightgray}{RGB}{248,248,248}
\definecolor{mydarkgray}{RGB}{83,83,83}
\definecolor{mypurple}{RGB}{102,57,186}
\definecolor{mygreen}{RGB}{0,119,0}
\definecolor{myblue}{RGB}{0,119,187}

\newcommand{\benchmark}{{{i}}PrefBench\xspace}
\newcommand{\eventnum}{\num{3051}\xspace}
\newcommand{\tasknum}{\num{130}\xspace}
\newcommand{\username}{\textit{Jordan Avery}\xspace}

\usepackage{booktabs}
\usepackage{pifont}
\usepackage{multirow}

\usepackage[dvipsnames]{xcolor}
\newcommand{\cmark}{\textcolor{ForestGreen}{\ding{51}}}
\newcommand{\xmark}{\textcolor{BrickRed}{\ding{55}}}
\newcommand{\pmark}{\textcolor{Orange}{$\ast$}}

\title{\benchmark: A User-Centric Benchmark for Personalized Mobile Use Agents}

\author[1, \dagger]{Yuting Ning}
\author[2]{Ruijia Cheng}
\author[2]{Sushobhan Ghosh}
\author[2]{Peitong Duan}
\author[2]{HungShi Lin}
\author[2]{Abhishek Sundararajan}
\author[2]{Shreyash Pandey}
\author[1]{Huan Sun}
\author[2]{Jeffrey Nichols}
\author[2]{Eldon Schoop}

\affiliation[1]{The Ohio State University}
\affiliation[2]{Apple}

\contribution{$^\dagger$ Work done during an internship at Apple.}

\abstract{
An agent that truly knows and acts on behalf of its user has been a long-standing promise of AI assistants. Computer-use agents bring this promise within reach by capably executing fully specified tasks across apps. However, real-world user requests are often underspecified. To act correctly for the particular user, an agent must \textit{infer} some missing information from the user's activity history and \textit{ask} for other information that only the user can provide, without making unsupported assumptions or burdening the user with unnecessary questions.
We present \benchmark, a benchmark for evaluating the \textit{infer} and \textit{ask} capabilities of personal mobile-use agents in an executable iOS environment with an interactive simulated user. Grounded in a coherent persona with \eventnum past activity events, \benchmark comprises \tasknum underspecified user tasks across \num{26} apps.
Each omitted user-dependent requirement is, by construction, either history-inferable or ask-only.
\benchmark evaluates personalized mobile agents on both whether agents complete tasks correctly for the user (i.e., personalized task completion) and whether they seek user input when necessary (i.e., user-agent interaction).
Evaluation of \num{7} frontier agents reveals a substantial gap between execution and personalization. Agents achieve high action completion scores (up to \num{91.6}\%) but low full task success (up to \num{47.7}\%) because personalization lags behind. They especially struggle with ask-only information, failing to request enough necessary information. Our findings highlight a challenge for personalized agents in knowing what to infer and when to ask.
We hope that \benchmark can serve as a valuable testbed for building agents that know their user and work with them.
}

\metadata[Correspondence]{\sffamily Yuting Ning: \url{ning.151@osu.edu}; Jeffrey Nichols: \url{jwnichols@apple.com}; Eldon Schoop: \url{eldon@apple.com}}
\date{\sffamily\today}

\begin{document}

\maketitle

\section{Introduction}

It has been a long-standing goal in Artificial Intelligence and Human-Computer Interaction to develop agents that can understand and act on behalf of users~\citep{maes1993learning, liebermanAgents,horvitz1999principles}. Recent computer-use agents (CUAs) have made rapid progress in completing digital tasks across apps and over long horizons~\citep{osworld, yuan2026osworld2, androidworld, ye2025mobile}. These agents often excel at fully specified tasks (e.g., ``make a reservation at Harborline Seafood tomorrow at 7:30 PM for a party of four.''). However, realistic user queries rarely provide complete details, and are often underspecified~\citep{volkel2021eliciting}. For example, a user may say ``book me my usual restaurant for tomorrow'' instead of specifying the restaurant name, reservation time, and party size.

\begin{figure*}
    \centering
    \includegraphics[width=1\linewidth]{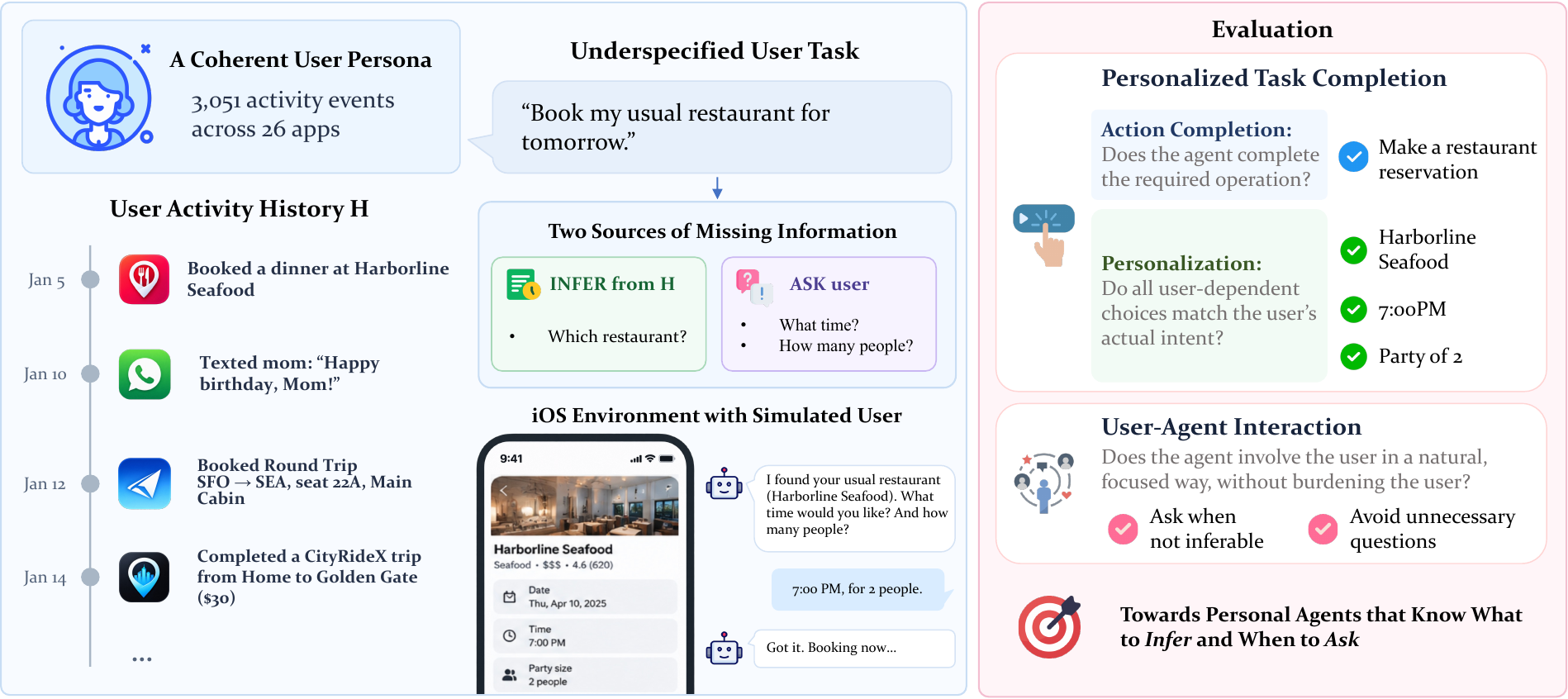}
    \caption{\benchmark evaluates personalized mobile-use agents on underspecified user tasks that require both \textit{inferring} user-specific information from grounded activity history and \textit{asking} the user for information unavailable from history.}
    \label{fig:teaser}
\end{figure*}

Contending with such underspecified tasks requires \emph{personalization} capabilities, in which an agent resolves missing information in two ways: \emph{infer} from knowledge of the user (e.g., the user's usual restaurant from past reservations and orders), and \textit{ask} for details that the available history does not establish (e.g., the desired time and party size).
An agent that silently makes unsupported assumptions can diverge from a user's actual intent, while an agent that asks for details already inferable from the user's history defeats the purpose of autonomy.
A truly useful personal agent must reason not only about what information is missing, but also where it should come from.

Yet this challenge is not fully captured by existing evaluations.
Typical CUA benchmarks fully specify every task and cold-start agents without any knowledge of the user~\citep{osworld, androidworld, webarena}, leaving nothing to infer or ask.
Benchmarks with underspecified tasks usually supply only one resolution source: personalization benchmarks expose a user history but offer no one to ask~\citep{lyu2026personalalign, yang2025fingertip}, while interactive benchmarks provide a simulated user but no user history that would make some questions unnecessary~\citep{yao2024tau, vijayvargiya2026ambig}. Some provide both~\citep{knowUBench,chen2026vitabench}, but do not distinguish which missing details should be inferred and which require clarification. Therefore, they cannot measure whether the agent knows what to infer and when to ask.

To bridge the gap, we present \benchmark, a benchmark designed around the infer-and-ask capabilities of personalized mobile-use agents in an executable iOS environment.
As shown in~\autoref{fig:teaser}, agents can know about the user through two sources: an activity history, provided as their only prior context, and a simulated user, reachable through an \texttt{ask\_user} action, who answers clarification questions. Both are grounded in rich seeded data around one coherent persona in the environment.
\benchmark comprises \tasknum underspecified tasks, and each omitted detail is designed to be either \emph{inferable} (i.e., grounded in specific evidence in activity history) or \emph{ask-only} (i.e., held solely by the simulated user).
To further assess whether agents can apply historical preferences appropriately in challenging real-world scenarios, we include tasks in which current instructions override past patterns, recent behavior indicates changing preferences, or preferred options are unavailable.

\benchmark assesses both whether agents complete tasks correctly for the user (i.e., \textit{personalized task completion}) and whether they seek user input when necessary (i.e., \textit{user-agent interaction}).
Personalized task completion is measured by task-specific rubrics, split into two types: \textit{action-completion} criteria measure whether required operations were successfully executed (e.g., a restaurant reservation was made), while \textit{personalization} criteria measure whether each user-dependent decision matches the intended values (e.g., the restaurant, time, and party size).
However, correct outcomes alone do not establish user satisfaction: an agent may succeed by interrogating the user about every detail, or luckily guessing the ask-only information correctly.
Therefore, inspired by Grice's maxims~\citep{grice1975logic}, we also evaluate user-agent interaction through recall and precision on ask-only information: recall measures whether the agent elicits all required ask-only information, while precision measures whether its requests obtain new ask-only information rather than ask for inferable or irrelevant details or repeat earlier requests.

We evaluate seven frontier agents on \benchmark. The results reveal a substantial gap between executing a task successfully and personalizing it for a particular user. Although Claude Opus 5, the best-performing agent evaluated, achieves a \num{91.6}\% action-completion score, its task success rate is only \num{47.7}\% because personalization lags behind.
Agents struggle particularly with ask-only information, and 
our interaction evaluation reveals a corresponding shortfall.
By default, agents ask far too little: the recall on ask-only information is low across the board, with the highest at only \num{44.3}\%.
Yet what they do ask is not reliably necessary either, with precision between \num{60}\% and \num{77}\%.
Explicitly adding a clarification phase can substantially increase the number of questions and raise the recall and task success rate, but lower precision in turn, exposing a tradeoff between silent guessing and excessive user burden.
The challenge is therefore not simply to ask more, but to determine when asking is necessary.
We hope \benchmark can serve as a valuable testbed for building agents that know their user and work with them effectively.

\section{Related Work}

\textbf{Computer-Use Agent Benchmarks.}
Computer-use agents have advanced rapidly~\citep{zheng2024gpt,hong2024cogagent,qin2025ui,wang2026opencua}, supported by executable benchmarks across desktop~\citep{osworld, yuan2026osworld2}, mobile~\citep{androidworld}, and web environments~\citep{webarena, koh2024visualwebarena, xue2025illusion, he2024webvoyager}.
These benchmarks typically use fully specified tasks without prior user knowledge.
MobileWorld~\citep{kong2026mobileworld} introduces underspecified tasks and an \texttt{ask\_user} action, but without user history, every missing detail must be asked.

\noindent\textbf{Personalized Agent Benchmarks.}
A growing line of work evaluates preference inference from user histories for personalized tasks, but does not support interactive clarification~\citep{cai2025large,yang2025fingertip, wang2026me, kim2026persona2web, lyu2026personalalign}.
iOSWorld~\citep{iosworld}, on which our environment builds, seeds an iOS simulator with a coherent persona but neither exposes user history as prior context nor supports user interaction.
VitaBench~2.0~\citep{chen2026vitabench} and KnowU-Bench~\citep{knowUBench} incorporate user histories and, in some cases, hidden information accessible through simulated users, but do not explicitly distinguish whether each missing detail is recoverable from the history or known only to the user, and thus cannot evaluate the efficiency of user-agent interaction, i.e., whether agents only ask necessary questions.

\noindent\textbf{Interactive and Clarification Benchmarks.}
Interactive benchmarks~\citep{yao2024tau,qian2024tell,qian2025userbench,barres2025tau2bench,vijayvargiya2026ambig} study clarification with simulated users whom agents have no prior knowledge of, and mainly measure task outcomes or how much hidden information agents elicit.
Ask-before-Plan~\citep{askbeforeplan} penalizes missed and unnecessary questions, while AppWorld-UL~\citep{chen2026appworld} evaluates information elicitation under underspecification, infeasibility, and confirmation-seeking scenarios.
However, all of these settings lack user histories against which the necessity of clarification questions can be assessed.

In contrast, \benchmark evaluates agents on underspecified tasks, whose missing information can either be inferred from grounded user history or only be obtained by asking the user, and explicitly distinguishes the resolution source, enabling evaluation of personalized task completion and necessity-aware user-agent interaction. We provide a detailed comparison in \autoref{tab:related_work}.
\section{\benchmark}

\subsection{Benchmark Overview}
\label{subsec:overview}

\benchmark evaluates whether mobile-use agents can complete underspecified tasks for a user, by inferring missing information from user history and seeking clarification when necessary.

For each task $t\in \mathcal{T}$, let $U_t$ denote the user-dependent details left unresolved by the instruction. Each $u \in U_t$ has an intended value and exactly one resolution source:
(1) \textbf{Inferable information} ($u \in U_t^{\mathrm{infer}}$): Information whose intended value can be recovered from consistent evidence in the user activity history $\mathcal{H}$, such as a dominant pattern across past events or a stable user fact.
(2) \textbf{Ask-only information} ($u \in U_t^{\mathrm{ask}}$): Information unavailable in $\mathcal{H}$ that must be obtained by asking the user, such as the context of the current occasion.

This split mirrors what a personal assistant can know about its user: stable preferences and habits leave traces in past activity, whereas the specifics of the current situation do not.
This distinction allows us to evaluate both whether agents infer user-specific details correctly from the history, and whether they seek clarification only when needed, without unnecessary questions.

\subsection{Environment}
\label{subsec:env}

\textbf{Executable iOS Environment.}
\benchmark builds on the executable iOS simulator from iOSWorld~\citep{iosworld}, containing \num{26} mock apps populated with interconnected data (e.g., past restaurant reservations, financial transactions, messages, workouts and ticket orders) for a coherent user persona, \username.
The agent operates through standard mobile GUI actions (e.g., \texttt{tap}, \texttt{type}, and \texttt{swipe}).
We extend the action space with an \texttt{ask\_user} action, which allows the agent to send a natural language message to the simulated user and receive a user response. Full details of the action space are provided in Appendix~\ref{app:action_space}.

\noindent\textbf{User Activity History.}
We compose seeded application data as a chronological user activity history $\mathcal{H}$ (Appendix~\ref{app:user_log}), representing a timeline of what the user did on their personal phone. Each event is converted from the underlying application metadata into natural-language descriptions. The activity history $\mathcal{H}$ is provided to the agent as the only prior knowledge and context of the user. All inferable information (e.g., recurring preferences, habits and user facts) is designed to be recoverable from evidence in $\mathcal{H}$.

\noindent\textbf{User Profile and User Simulator.}
To ground simulated user responses and facilitate evaluation, we derive a structured user profile $\mathcal{P}$ from the seeded application data (Appendix~\ref{app:user_profile}), consolidating the persona's preferences, habits, and personal information into \num{13} dimensions (e.g., identity, social relationships, finances, dining and travel preferences, and fitness habits).
$\mathcal{P}$ is hidden from the agent and serves as a ground-truth reference for user simulation and evaluation.
We simulate the interactive user with an LLM grounded in $\mathcal{P}$ and, when applicable, task-specific hidden context. When the agent issues an \texttt{ask\_user} action, the user simulator answers the question in character.
To faithfully evaluate the agent's own ability to recognize what is missing, we instruct the simulator to answer only what is asked, without proactively volunteering additional information.

\subsection{Tasks}
\label{subsec:tasks}

\textbf{Task Design.}
As described in \S\ref{subsec:overview}, each task intentionally leaves user-dependent details unspecified, with their intended values and resolution sources explicitly annotated.
For example, ``book me my usual restaurant for tomorrow'' leaves three pieces of information underspecified: the restaurant, party size, and time. The user's reservation history in $\mathcal{H}$ can resolve the usual restaurant, so asking the user about it is redundant. In contrast, the reservation time and party size, which describe the current occasion and cannot be supported by any history, must be obtained by asking rather than assumed.
A single task may contain multiple inferable and ask-only requirements, and often requires agents to integrate supporting evidence across multiple historical events or transfer preferences inferred in one app to decisions in another. Success therefore requires reasoning about the user rather than simply retrieving a single past event.

Real-world personalization can be even more complex, as things can diverge from past trends and historical behavior cannot be treated as unconditional specification of future intent. Therefore, we introduce three perturbation types:
(1) \textit{Instruction perturbation}: The current user instruction explicitly contradicts the dominant historical pattern. For example, a user who typically selects an inexpensive tier for rideshare may explicitly ask for a premium ride for a special trip. The agent must prioritize the current instruction over an inferred historical preference.
(2) \textit{Preference perturbation}: Recent behavior in $\mathcal{H}$ suggests that the user's preference has shifted over time and departed from the long-run mode.
For example, the user selected aisle seats consistently in recent months, while most historical flights were window seats. A personalized agent should therefore identify the evolving preference trend, rather than simply selecting the most frequent historical value.
(3) \textit{Environment perturbation}: The preferred option becomes unavailable in the current environment. The agent should respond to the situation by recognizing the unavailability, surfacing it to the user, and acting according to the user's preference for alternatives or stopping if no substitution is acceptable.

\noindent\textbf{Task Authoring and Quality Assurance.}
We use a three-stage task authoring and validation pipeline to ensure realistic underspecification, task feasibility, grounded resolution for each omitted requirement, and verifiable outcomes.
First, coding agents with \textit{task-authoring skills} generate candidate personalized tasks using the source code of the apps, user activity history and profile, and fully specified iOSWorld tasks as references.
Second, human \textit{task-refinement experts} review and revise each candidate task.
Specifically, for each omitted detail, experts verify its resolution source.
An inferable detail must be supported by a dominant pattern or explicit fact in $\mathcal{H}$, and where the evidence is weak or mixed, we revise the seeded data so that $\mathcal{H}$ can support a clear preference sufficiently. An ask-only detail must not be recoverable from $\mathcal{H}$, and its intended value is chosen to differ from the app's pre-filled default when possible, so that silently accepting a default or guessing is unlikely to succeed by chance.
At this stage, experts also remove duplicate tasks and diversify the candidate tasks to cover as many user preferences and apps as possible.
Finally, \textit{experienced validation experts} execute refined tasks in the simulator to verify environmental feasibility and rubric correctness, and correct remaining issues accordingly.
Every task in the final \benchmark therefore undergoes two rounds of human review and revision.
More details are provided in Appendix~\ref{app:task_collection}.

\subsection{Evaluation}
\label{subsec:eval}
A personalized agent should complete user requests successfully with appropriate personalization, asking the user for additional information only when necessary. 
There are multiple failure modes: it may execute the correct operation while making wrong user-dependent choices (e.g., purchasing a game controller as requested, but not the model compatible with the user's console), or eventually reach the correct outcome by repeatedly asking the user for every detail.
We therefore evaluate two aspects of agent behavior: (1) Personalized Task Completion, for whether the agent \emph{completes} the required operations and whether it \emph{personalizes} them for the user correctly; and (2) User-Agent Interaction, for whether it involves the user only when necessary. More details are provided in Appendix~\ref{app:judge}.

\noindent\textbf{Personalized Task Completion.}
Each task is accompanied by a task-specific rubric containing two types of criteria.
Each \textbf{action-completion} criterion evaluates an operation required by the instruction, independent of user-specific choices. For instance, a criterion requires that the agent successfully creates a restaurant reservation, regardless of the restaurant, time, or party size.
Each \textbf{personalization} criterion corresponds to a user-dependent decision involved in the task, and is additionally labeled with its resolution source (i.e., inferable or ask-only).
For example, correctly reserving the user's usual restaurant satisfies an inferable personalization criterion, and reserving at the intended time and party size satisfies the ask-only personalization criteria.
An LLM judge~\citep{zheng2023judging} evaluates each criterion as pass or fail using the execution trajectory (including screenshots, actions, and the final answer), user profile, task-specific hidden context, and rubric.
Action-completion and personalization scores are computed as the fractions of satisfied criteria in their corresponding category, macro-averaged over tasks. A task counts as a \emph{success} only if all of its criteria are satisfied.

\noindent\textbf{User-Agent Interaction.}
Following the cooperative principle in conversation that a contribution should be as informative as required, and no more~\citep{grice1975logic}, we evaluate whether agents request all ask-only information while avoiding unnecessary questions.
We first use an LLM judge to decompose \texttt{ask\_user} messages from the agent into a multiset of individual information requests $\mathcal{Q}_t$.
Multiple independent requests within one message are treated separately, and repeated requests are counted as separate occurrences.
Each request $q \in \mathcal{Q}_t$ is semantically matched to an ask-only requirement, if any:
$\phi_t(q) \in U_t^{\mathrm{ask}} \cup \{\bot\}$,
where $\bot$ denotes no match.
The set of distinct ask-only requirements elicited by the agent is
$
E_t = \left\{
u \in U_t^{\mathrm{ask}}
\;\middle|\;
\exists q \in \mathcal{Q}_t,\ \phi_t(q)=u
\right\}.
$
We then compute interaction Recall and Precision as
$\text{Recall}=\sum_t|E_t|/\sum_t|U^{ask}_t|$ and $\text{Precision}=\sum_t|E_t|/\sum_t|Q_t|$.
Recall measures the fraction of ask-only details the agent elicits, while Precision measures the fraction of request occurrences that cover a new ask-only requirement, penalizing requests for inferable or irrelevant information as well as repeated requests.
Together, these metrics penalize both agents that silently fabricate what they cannot know and those that excessively burden the user with exhaustive questions.

\subsection{Benchmark Statistics}

The resulting \benchmark contains \tasknum tasks, grounded in \eventnum user activity events across \num{26} apps. Among these tasks, \num{58.5}\% contain at least one ask-only requirement, and \num{26.9}\% require cross-app preference transfer. The benchmark also includes \num{11} instruction-perturbation, \num{10} preference-perturbation, and \num{13} environment-perturbation tasks.
Each task contains an average of \num{2.09} action-completion and \num{2.87} personalization criteria, of which \num{1.98} are inferable and \num{0.89} are ask-only. More statistics are in Appendix~\ref{app:statistic}.
\section{Experiments}
\label{sec:exp}

\begin{table*}[t]
\centering
\small
\resizebox{\textwidth}{!}{%
\begin{tabular}{lccccccccc}
\toprule
\multirow{3}{*}{\textbf{Agent}}
& \multicolumn{5}{c}{\textbf{Personalized Task Completion}}
& \multicolumn{3}{c}{\textbf{User-Agent Interaction}}
& \multirow{3}{*}{\textbf{\#Steps}} \\
\cmidrule(lr){2-6} \cmidrule(lr){7-9}
& \multirow{2}{*}{SR}
& \multirow{2}{*}{Action}
& \multicolumn{3}{c}{Personalization}
& \multirow{2}{*}{Recall}
& \multirow{2}{*}{Precision}
& \multirow{2}{*}{\#Ask}
& \\
\cmidrule(lr){4-6}
& & & Overall & Infer & Ask & & & & \\
\midrule
Claude Opus 5
& \num{47.7} & \num{91.6} & \num{69.4}
& \num{76.3} & \num{51.3}
& \num{44.2} & \num{65.8} & \num{0.35}
& \num{22.5} \\

Claude Sonnet 5
& \num{39.2} & \num{83.1} & \num{64.8}
& \num{74.2} & \num{41.5}
& \num{25.7} & \num{60.4} & \num{0.24}
& \num{22.0} \\

Gemini 3.1 Pro
& \num{36.2} & \num{78.6} & \num{60.9}
& \num{69.2} & \num{40.1}
& \num{23.9} & \num{77.1} & \num{0.20}
& \num{22.6} \\

Gemini 3.7 Flash
& \num{30.0} & \num{79.6} & \num{55.1}
& \num{67.5} & \num{21.1}
& \num{10.6} & \num{70.6} & \num{0.08}
& \num{25.6} \\

Kimi K3
& \num{36.2} & \num{85.8} & \num{63.9}
& \num{73.0} & \num{39.3}
& \num{15.9} & \num{75.0} & \num{0.15}
& \num{23.1} \\

Qwen3.8-27B
& \num{26.2} & \num{77.5} & \num{54.4}
& \num{62.1} & \num{29.4}
& \num{21.2} & \num{70.6} & \num{0.21}
& \num{20.7} \\

Qwen3.8-Flash-Next
& \num{31.5} & \num{77.6} & \num{58.6}
& \num{67.8} & \num{33.2}
& \num{25.7} & \num{60.4} & \num{0.26}
& \num{24.7} \\
\bottomrule
\end{tabular}%
}
\caption{Main results on \benchmark. Personalized task completion is measured by success rate (SR), action completion and personalization score. We further divide the personalization score by information resolution source (i.e., inferable and ask-only). For user-agent interaction, we report Recall, Precision, and the average number of \texttt{ask\_user} calls per task (\#Ask). \#Steps denotes the average number of actions per task. All scores are percentages except \#Ask and \#Steps.}
\label{tab:main}
\end{table*}

\subsection{Experimental Setup}

We evaluate seven frontier models under the same mobile-use agent harness on \benchmark: Claude Opus 5~\citep{claude_opus5}, Claude Sonnet 5~\citep{claude_sonnet5}, Gemini 3.1 Pro~\citep{google_gemini3_1_pro}, Gemini 3.7 Flash~\citep{google_gemini3_7_flash}, Kimi K3~\citep{team2026kimi}, Qwen3.8-27B~\citep{qwen38}, and Qwen3.8-Flash-Next~\citep{qwen3.8flashnext}.
We also attempted to evaluate specialized GUI agents, but their limited context capacity or frequent invalid actions under our setting prevented meaningful personalization comparisons (Appendix~\ref{app:agents}).
Each episode runs on a freshly seeded iOS simulator, with task-specific perturbations applied when applicable. Agents receive the full user activity log as context in all experiments except the study on user history exposure, to avoid confounding preference inference with retrieval or navigation ability. At each step, the agent observes the current screenshot and acts through the standard mobile-use action space including \texttt{ask\_user} (Appendix~\ref{app:action_space}). Episodes end when the agent issues a \texttt{stop} action or reaches the maximum \num{50}-step budget. We use Claude Sonnet 5 for both the user simulator and LLM judges (validated in \S\ref{subsec:validation}).
Following \S\ref{subsec:eval}, we report \emph{full task Success Rate (SR)}, \emph{Action Completion} and \emph{Personalization} scores, and \emph{interaction Recall} and \emph{Precision}. Further details of the experimental setup are in Appendix~\ref{app:exp}.

\subsection{Main Results}
\autoref{tab:main} shows the performance of evaluated agents on \benchmark, with the following observations:

\begin{figure*}
    \centering
    \includegraphics[width=0.95\linewidth]{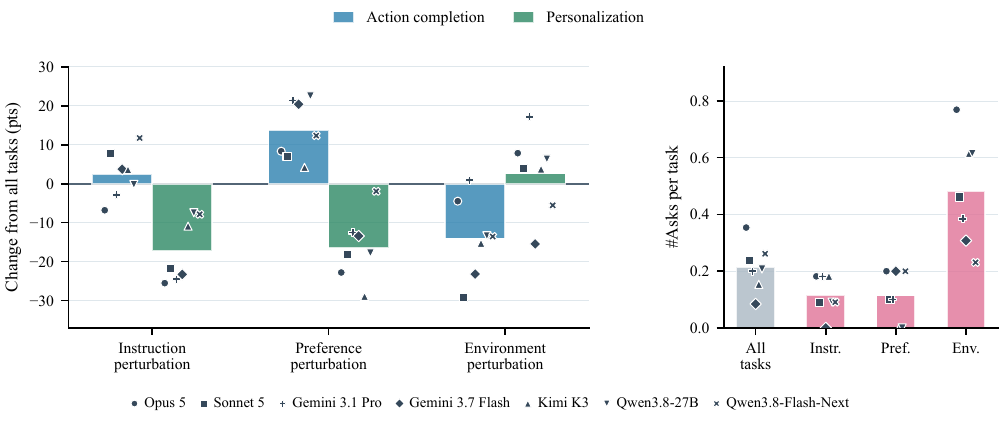}
    \caption{Performance on the three perturbation subsets of \benchmark. \textbf{Left}: action completion and personalization differences, computed for each agent as its subset score minus its overall score on the full benchmark.
    Bars are means across agents and markers are individual agents. \textbf{Right}: average number of \texttt{ask\_user} actions per task.}
    \label{fig:perturbation}
\end{figure*}

\noindent\textbf{Strong execution, weak personalization.} 
Across all agents, action execution is substantially stronger than personalization.
While agents achieve relatively high action-completion scores (\num{77.5}\% --\num{91.6}\%), the full task success drops to \num{26.2}\%--\num{47.7}\% because personalization lags behind. For example, Claude Opus 5, the best-performing agent we evaluate, completes \num{91.6}\% of required actions but only \num{69.4}\% of user-dependent decisions and therefore succeeds on only \num{47.7}\% of tasks.
These gaps show that successfully executing required operations does not necessarily lead to user satisfaction. Frontier agents that can excel at short- to medium-horizon mobile navigation still fail to resolve user-dependent decisions for a specific user.

\noindent\textbf{Ask-only information poses a particular challenge: agents struggle to decide when to ask.}
Splitting personalization by resolution source, performance on ask-only information is consistently lower than on inferable information across the board, by up to \num{46} points for Gemini 3.7 Flash.
This suggests that recognizing and resolving information that requires user input during execution remains a major bottleneck for current agents.
The interaction metrics locate the cause in how agents involve the user. Although tasks contain \num{0.89} ask-only requirements on average, agents issue only \num{0.08} to \num{0.35} \texttt{ask\_user} calls per task, eliciting at most \num{44.3}\% of the information they could not know (i.e., Recall). Meanwhile, what they do ask is not reliably necessary either. Precision ranges from \num{60.4}\% to \num{77.1}\%, showing that a substantial fraction of their information requests are unnecessary or redundant.
Agents thus struggle not merely with asking too little, but with identifying when asking is necessary.

\noindent\textbf{Personalization is sensitive to conflicts with historical preferences.}
We further examine agent performance on the three perturbation subsets in \autoref{fig:perturbation}.
On instruction- and preference-perturbation subsets, personalization scores are lower than the overall benchmark averages, whereas action-completion scores do not show a comparable decline.
In contrast, environment perturbations encourage agents to ask more. When the historically preferred option becomes unavailable, agents start to seek user input, issuing \num{2.3}$\times$ as many questions, and personalization on ask-only information improves for every agent. This contrast suggests that agents are more responsive to explicit environmental obstacles than to uncertainty that must be recognized from context by themselves.

\subsection{Error Analysis}
\label{subsec:error_analysis}

\begin{figure*}
    \centering
    \includegraphics[width=0.95\linewidth]{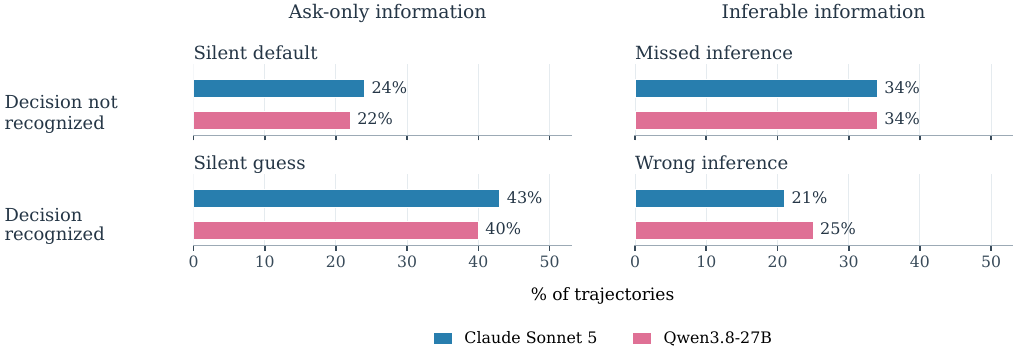}
    \caption{Distribution of personalization failures along two dimensions: whether the unresolved information is ask-only or inferable, and whether the agent recognizes the corresponding decision. A trajectory may contain multiple errors.}
    \label{fig:error_analysis}
\end{figure*}

To better understand where personalization fails, we manually label failed trajectories from Claude Sonnet 5 and Qwen3.8-27B.
We categorize personalization errors along two dimensions:  \emph{where} the missing information must come from (user history or the user) and whether the agent recognized \emph{what} information was needed.
The categorization yields four error types in~\autoref{fig:error_analysis}.

For ask-only information, agents may recognize that a value is unspecified but nevertheless commit to one with their own guess without consulting the user (i.e., \textit{Silent Guess}, Ex.~\ref{fig:silent_guess}). Examining Qwen3.8-27B's reasoning, we find that it often raises ``should I ask?'' before talking itself into a guess or over-reasoning from past behavior.
In other cases, the missing requirements are never recognized as an open decision at all (\textit{Silent Default}, Ex.~\ref{fig:silent_default}).
For example, an agent asked to book a round trip without specified travel dates never considers the dates to be user preferences at all, and the app's pre-filled dates simply pass through unchanged.
Inferable information exhibits an analogous distinction. Some preferences are never surfaced as decisions (\textit{Missed Inference}, Ex.~\ref{fig:missed_inference}), while others are recognized but resolved incorrectly from history (\textit{Wrong Inference}, Ex.~\ref{fig:wrong_inference}).

These errors suggest two distinct challenges for current agents: (1) identifying which user-dependent choices remain unresolved, and (2) once a choice is recognized, determining whether its value should be inferred from history or elicited from the user, and resolving it correctly.
Detailed definitions and examples of these error types are provided in Appendix~\ref{app:error_analysis}, with additional qualitative analysis and case studies in Appendix~\ref{app:add_case_studies}.

\subsection{Additional Analysis and Discussion}
\label{sec:analysis}

To explore how agent personalization might be improved, we study two factors that correspond to two sources of knowledge about the user: explicitly prompting the agent to clarify unresolved
information before execution, and varying how much of the user history is exposed to the agent.

\begin{table*}[t]
\centering
\small
\resizebox{0.95\textwidth}{!}{%
\begin{tabular}{llccccccccr}
\toprule
\multirow{3}{*}{\textbf{Agent}} & \multirow{3}{*}{\textbf{Clarification}}
& \multicolumn{5}{c}{\textbf{Personalized Task Completion}}
& \multicolumn{3}{c}{\textbf{User-Agent Interaction}}
& \multirow{3}{*}{\textbf{\#Steps}} \\
\cmidrule(lr){3-7} \cmidrule(lr){8-10}
& & \multirow{2}{*}{SR}
& \multirow{2}{*}{Action}
& \multicolumn{3}{c}{Personalization}
& \multirow{2}{*}{Recall}
& \multirow{2}{*}{Precision}
& \multirow{2}{*}{\#Ask}
& \\
\cmidrule(lr){5-7}
& & & & Overall & Infer & Ask & & & & \\
\midrule
\multirow{2}{*}{Claude Sonnet 5}
& on-demand & \num{39.2} & \num{83.1} & \num{64.8} & \num{74.2} & \num{41.5} & \num{25.7} & \num{60.4} & \num{0.24} & \num{22.0} \\
& \quad + pre-execution & \num{53.8} & \num{82.0} & \num{77.8} & \num{73.9} & \num{82.0} & \num{85.0} & \num{42.5} & \num{1.37} & \num{24.5} \\
\midrule
\multirow{2}{*}{Qwen3.8-27B}
& on-demand & \num{26.2} & \num{77.5} & \num{54.4} & \num{62.1} & \num{29.4} & \num{21.2} & \num{70.6} & \num{0.21} & \num{20.7} \\
& \quad + pre-execution & \num{35.4} & \num{76.6} & \num{67.8} & \num{66.0} & \num{65.0} & \num{69.0} & \num{60.9} & \num{0.84} & \num{21.5} \\
\bottomrule
\end{tabular}
}
\caption{Effect of an explicit pre-execution clarification phase on Claude Sonnet 5 and Qwen3.8-27B. All scores are percentages except \#Ask and \#Steps.}
\label{tab:clarify}
\vspace{-1em}
\end{table*}

\begin{wrapfigure}[19]{r}{0.5\linewidth}
\vspace{-1.5\baselineskip}
\centering
\includegraphics{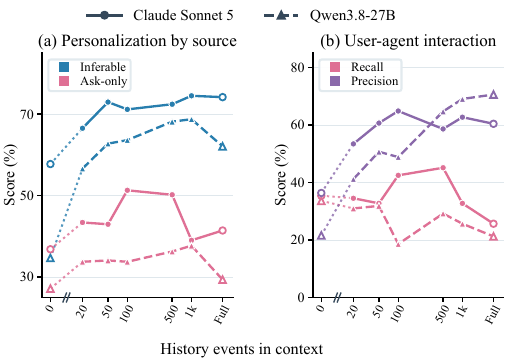}
\caption{Effect of different amounts of user history exposed in the agent's context. (a) Personalization score split by information source. (b) User-agent interaction measured by recall and precision.}
\label{fig:memory}
\end{wrapfigure}

\noindent\textbf{Effect of Explicit Clarification.}
Motivated by the low interaction recall observed, we introduce
a pre-execution clarification phase, in which the agent examines
the user task and activity history, identifies ambiguities that
cannot be resolved from history, and asks the user before
executing the task.
As shown in~\autoref{tab:clarify}, the explicit clarification phase substantially improves personalization and interaction recall, eliciting more questions from the agent. However, interaction precision decreases at the same time, indicating that explicit clarification alleviates under-asking at the cost of more unnecessary requests. Simply encouraging agents to seek clarification improves coverage of ask-only requirements, but does not help agents distinguish necessary requests from unnecessary ones.

\noindent\textbf{Effect of User History Exposure.}
We further investigate how performance changes when different user history is exposed to the agent, ranging from no history to different retrieved subsets of task-relevant events, and the full unfiltered history. Task-relevant events are retrieved using the task instruction as query and text-embedding-3-small~\citep{openai2024embedding} as the embedder.
As shown in \autoref{fig:memory}(a), inferable personalization rises sharply with a small set of relevant events (e.g., from \num{35}\% to \num{63}\% for Qwen3.8-27B with the top-50 events) and then plateaus, with the full history adding no further gain.
Ask-only personalization, in contrast, barely moves.
\autoref{fig:memory}(b) shows how history exposure instead changes user-agent interaction: with more history, agents ask fewer questions and interaction precision rises, as questions whose answers are already in the history are avoided.
However, this does not necessarily indicate more appropriate interaction, as recall on ask-only information does not improve and even decreases (e.g., $33.6\rightarrow21.2$ for Qwen3.8-27B).
More user history thus helps agents infer preferences and avoid unnecessary questions, but does not help them recognize what genuinely remains unresolved.

\subsection{Validation of User Simulator and LLM Judges}
\label{subsec:validation}

We validate whether the user simulator faithfully answers agents' questions and whether the LLM judges reliably produce assessments.

\noindent\textbf{User Simulator.}
We inspect all question–answer pairs produced when interacting with four representative agents, and \num{96.6}\% of responses are grounded in the user profile or task-specific hidden context and answer exactly what was asked without volunteering additional information (Appendix~\ref{app:user_sim_validation}).

\noindent\textbf{LLM Judges.}
We first validate the LLM-based evaluation against human annotations on \num{100} sampled trajectories. Human annotators label each task-rubric criterion and each message issued by the agent.
Our LLM judges agree with human labels on \num{95.7}\% of rubric criteria ($\kappa = 0.89$), \num{97.0}\% of task-level success ($\kappa = 0.94$), and \num{95.2}\% / \num{93.9}\% of recall / precision judgments ($\kappa = 0.90$ / $0.82$), and results are consistent with Gemini 3.7 Flash as an alternative judge (Appendix~\ref{app:human_validation}).
\section{Conclusion}

In this work, we introduce \benchmark, a user-centric benchmark for evaluating personalized agents in \tasknum underspecified mobile-use tasks. We center on two core capabilities for personalized agents: inferring user preferences from the user's activity history, and asking the user for clarification when necessary. For each underspecified task, the agent needs to determine not only which information is missing, but also where it should come from.
By controlling the provenance of each omitted detail, \benchmark enables evaluation of whether the agent can execute the required action (i.e., action completion) and complete the task correctly for the particular user (i.e., personalization), and whether it involves the user only when necessary (i.e., user-agent interaction).
Our evaluation on seven frontier agents reveals a clear gap between action execution and personalization, especially when the user preferences can only be obtained by asking. Agents struggle most to recognize which information is genuinely unavailable and requires user input. Explicitly encouraging clarification increases the number of questions, but at the cost of over-asking. These findings suggest that we are still far from a truly personalized agent that can know what to infer and when to ask.

\section*{Acknowledgment}
The authors would like to thank Dongxu Li for valuable feedback and colleagues from the OSU NLP group for constructive discussion.

\clearpage




\applefootnote{ \textcolor{textgray}{\sffamily Apple and the Apple logo are trademarks of Apple Inc., registered in the U.S. and other countries and regions.}}

\bibliographystyle{plainnat}
\bibliography{custom}

@article{knowUBench,
  title={Knowu-bench: Towards interactive, proactive, and personalized mobile agent evaluation},
  author={Chen, Tongbo and Lu, Zhengxi and Xu, Zhan and Shao, Guocheng and Zhao, Shaohan and Tang, Fei and Du, Yong and Song, Kaitao and Liu, Yizhou and Yan, Yuchen and others},
  journal={arXiv preprint arXiv:2604.08455},
  year={2026}
}

@article{iosworld,
  title={iOSWorld: A benchmark for personally intelligent phone agents},
  author={Jang, Lawrence Keunho and Woodside, Mareks and Carom, Geronimo and Jang, Andrew Keunwoo and Koh, Jing Yu and Salakhutdinov, Ruslan},
  journal={arXiv preprint arXiv:2606.09764},
  year={2026}
}

@article{osworld,
  title={Osworld: Benchmarking multimodal agents for open-ended tasks in real computer environments},
  author={Xie, Tianbao and Zhang, Danyang and Chen, Jixuan and Li, Xiaochuan and Zhao, Siheng and Cao, Ruisheng and Hua, Toh J and Cheng, Zhoujun and Shin, Dongchan and Lei, Fangyu and others},
  journal={Advances in Neural Information Processing Systems},
  volume={37},
  pages={52040--52094},
  year={2024}
}

@inproceedings{androidworld,
  title={Androidworld: A dynamic benchmarking environment for autonomous agents},
  author={Rawles, Chris and Clinckemaillie, Sarah and Chang, Yifan and Waltz, Jonathan and Lau, Gabrielle and Fair, Marybeth and Li, Alice and Bishop, William and Li, Wei and Campbell-Ajala, Folawiyo and others},
  booktitle={International Conference on Learning Representations},
  volume={2025},
  pages={406--441},
  year={2025}
}

@inproceedings{webarena,
  title={Webarena: A realistic web environment for building autonomous agents},
  author={Zhou, Shuyan and Xu, Frank F and Zhu, Hao and Zhou, Xuhui and Lo, Robert and Sridhar, Abishek and Cheng, Xianyi and Ou, Tianyue and Bisk, Yonatan and Fried, Daniel and others},
  booktitle={International Conference on Learning Representations},
  volume={2024},
  pages={15585--15606},
  year={2024}
}

@article{barres2025tau2bench,
  title={$\tau^2$-Bench: Evaluating Conversational Agents in a Dual-Control Environment},
  author={Barres, Victor and Dong, Honghua and Ray, Soham and Si, Xujie and Narasimhan, Karthik},
  journal={arXiv preprint arXiv:2506.07982},
  year={2025}
}

@inproceedings{askbeforeplan,
  title={Ask-before-plan: Proactive language agents for real-world planning},
  author={Zhang, Xuan and Deng, Yang and Ren, Zifeng and Ng, See Kiong and Chua, Tat-Seng},
  booktitle={Findings of the association for computational linguistics: EMNLP 2024},
  pages={10836--10863},
  year={2024}
}

@inproceedings{maes1993learning,
    title={Learning Interface Agents},
    author={Pattie Maes and Robyn Kozierok},
    booktitle={Proceedings of the Eleventh National Conference on Artificial Intelligence (AAAI-93)},
    pages={459--465},
    year={1993},
    publisher={AAAI Press},
}

@inproceedings{liebermanAgents,
  title={Autonomous interface agents},
  author={Lieberman, Henry},
  booktitle={Proceedings of the ACM SIGCHI Conference on Human factors in computing systems},
  pages={67--74},
  year={1997}
}

@article{zhou2025maiUI,
  title={MAI-UI Technical Report: Real-World Centric Foundation GUI Agents},
  author={Zhou, Hanzhang and Zhang, Xu and Tong, Panrong and Zhang, Jianan and Chen, Liangyu and Kong, Quyu and Cai, Chenglin and Liu, Chen and Wang, Yue and Zhou, Jingren and others},
  journal={arXiv preprint arXiv:2512.22047},
  year={2025}
}

@article{grice1975logic,
  title={Logic and conversation},
  author={Grice, H Paul and Cole, Peter and Morgan, Jerry L and others},
  journal={1975},
  volume={41},
  pages={58},
  year={1975}
}

@inproceedings{horvitz1999principles,
  title={Principles of mixed-initiative user interfaces},
  author={Horvitz, Eric},
  booktitle={Proceedings of the SIGCHI conference on Human Factors in Computing Systems},
  pages={159--166},
  year={1999}
}

@inproceedings{vijayvargiya2026ambig,
  title={Ambig-swe: Interactive agents to overcome underspecificity in software engineering},
  author={Vijayvargiya, Sanidhya and Zhou, Xuhui and Yerukola, Akhila and Sap, Maarten and Neubig, Graham},
  booktitle={International Conference on Learning Representations},
  volume={2026},
  pages={146716--146737},
  year={2026}
}

@inproceedings{peng2025morae,
  title={Morae: Proactively pausing ui agents for user choices},
  author={Peng, Yi-Hao and Li, Dingzeyu and Bigham, Jeffrey P and Pavel, Amy},
  booktitle={Proceedings of the 38th Annual ACM Symposium on User Interface Software and Technology},
  pages={1--14},
  year={2025}
}

@article{zheng2023judging,
  title={Judging llm-as-a-judge with mt-bench and chatbot arena},
  author={Zheng, Lianmin and Chiang, Wei-Lin and Sheng, Ying and Zhuang, Siyuan and Wu, Zhanghao and Zhuang, Yonghao and Lin, Zi and Li, Zhuohan and Li, Dacheng and Xing, Eric and others},
  journal={Advances in neural information processing systems},
  volume={36},
  pages={46595--46623},
  year={2023}
}

@inproceedings{koh2024visualwebarena,
  title={Visualwebarena: Evaluating multimodal agents on realistic visual web tasks},
  author={Koh, Jing Yu and Lo, Robert and Jang, Lawrence and Duvvur, Vikram and Lim, Ming and Huang, Po-Yu and Neubig, Graham and Zhou, Shuyan and Salakhutdinov, Russ and Fried, Daniel},
  booktitle={Proceedings of the 62nd Annual Meeting of the Association for Computational Linguistics (Volume 1: Long Papers)},
  pages={881--905},
  year={2024}
}

@article{deng2023mind2web,
  title={Mind2web: Towards a generalist agent for the web},
  author={Deng, Xiang and Gu, Yu and Zheng, Boyuan and Chen, Shijie and Stevens, Sam and Wang, Boshi and Sun, Huan and Su, Yu},
  journal={Advances in Neural Information Processing Systems},
  volume={36},
  pages={28091--28114},
  year={2023}
}

@article{xue2025illusion,
  title={An illusion of progress? assessing the current state of web agents},
  author={Xue, Tianci and Qi, Weijian and Shi, Tianneng and Song, Chan Hee and Gou, Boyu and Song, Dawn and Sun, Huan and Su, Yu},
  journal={arXiv preprint arXiv:2504.01382},
  year={2025}
}

@inproceedings{kong2026mobileworld,
  title={Mobileworld: Benchmarking autonomous mobile agents in agent-user interactive and mcp-augmented environments},
  author={Kong, Quyu and Zhang, Xu and Yang, Zhenyu and Gao, Nolan and Liu, Chen and Tong, Panrong and Cai, Chenglin and Zhou, Hanzhang and Zhang, Jianan and Chen, Liangyu and others},
  booktitle={Proceedings of the 64th Annual Meeting of the Association for Computational Linguistics (Volume 1: Long Papers)},
  pages={6142--6167},
  year={2026}
}

@article{yuan2026osworld2,
  title={OSWorld2. 0: Benchmarking Computer Use Agents on Long-Horizon Real-World Tasks},
  author={Yuan, Mengqi and Zhou, Zilong and Xiong, Xinzhuang and Wu, Weiming and Sun, Jiayang and Song, Jiamin and Cui, Kaiqian and Wang, Bowen and Wu, Haoyuan and Li, Yitong and others},
  journal={arXiv preprint arXiv:2606.29537},
  year={2026}
}

@article{zheng2024gpt,
  title={Gpt-4v (ision) is a generalist web agent, if grounded},
  author={Zheng, Boyuan and Gou, Boyu and Kil, Jihyung and Sun, Huan and Su, Yu},
  journal={arXiv preprint arXiv:2401.01614},
  year={2024}
}

@inproceedings{hong2024cogagent,
  title={Cogagent: A visual language model for gui agents},
  author={Hong, Wenyi and Wang, Weihan and Lv, Qingsong and Xu, Jiazheng and Yu, Wenmeng and Ji, Junhui and Wang, Yan and Wang, Zihan and Dong, Yuxiao and Ding, Ming and others},
  booktitle={2024 IEEE/CVF Conference on Computer Vision and Pattern Recognition (CVPR)},
  pages={14281--14290},
  year={2024},
  organization={IEEE}
}

@article{wang2026opencua,
  title={Opencua: Open foundations for computer-use agents},
  author={Wang, Xinyuan and Wang, Bowen and Lu, Dunjie and Yang, Junlin and Xie, Tianbao and Wang, Junli and Deng, Jiaqi and Guo, Xiaole and Xu, Yiheng and Wu, Chen and others},
  journal={Advances in Neural Information Processing Systems},
  volume={38},
  pages={139756--139806},
  year={2026}
}

@article{qin2025ui,
  title={Ui-tars: Pioneering automated gui interaction with native agents},
  author={Qin, Yujia and Ye, Yining and Fang, Junjie and Wang, Haoming and Liang, Shihao and Tian, Shizuo and Zhang, Junda and Li, Jiahao and Li, Yunxin and Huang, Shijue and others},
  journal={arXiv preprint arXiv:2501.12326},
  year={2025}
}

@inproceedings{lyu2026personalalign,
  title={Personalalign: Hierarchical implicit intent alignment for personalized gui agent with long-term user-centric records},
  author={Lyu, Yibo and Chen, Gongwei and Shao, Rui and Guan, Weili and Nie, Liqiang},
  booktitle={Proceedings of the 64th Annual Meeting of the Association for Computational Linguistics (Volume 1: Long Papers)},
  pages={36074--36089},
  year={2026}
}

@inproceedings{wang2026me,
  title={Me-agent: A personalized mobile agent with two-level user habit learning for enhanced interaction},
  author={Wang, Shuoxin and Liu, Chang and Loo, Gowen and Zheng, Lifan and Wei, Kaiwen and Yan, Huanqian and Zeng, Xinyi and Zhang, Jingyuan and Tian, Yu},
  booktitle={Findings of the Association for Computational Linguistics: ACL 2026},
  pages={24206--24222},
  year={2026}
}

@article{yang2025fingertip,
  title={Fingertip 20k: A benchmark for proactive and personalized mobile llm agents},
  author={Yang, Qinglong and Li, Haoming and Zhao, Haotian and Yan, Xiaokai and Ding, Jingtao and Xu, Fengli and Li, Yong},
  journal={arXiv preprint arXiv:2507.21071},
  year={2025}
}

@article{yao2024tau,
  title={$tau$-bench: A Benchmark for Tool-Agent-User Interaction in Real-World Domains},
  author={Yao, Shunyu and Shinn, Noah and Razavi, Pedram and Narasimhan, Karthik},
  journal={arXiv preprint arXiv:2406.12045},
  year={2024}
}

@article{chen2026vitabench,
  title={VitaBench 2.0: Evaluating Personalized and Proactive Agents in Long-Term User Interactions},
  author={Chen, Yuxin and Zhang, Yi and Cai, Zhengzhou and Shi, Yaorui and Yao, Zhiyuan and Cui, Chenhang and Zheng, Jingnan and Huo, Yaqi and Su, Xi and Gu, Qi and others},
  journal={arXiv preprint arXiv:2605.27141},
  year={2026}
}

@inproceedings{he2024webvoyager,
  title={Webvoyager: Building an end-to-end web agent with large multimodal models},
  author={He, Hongliang and Yao, Wenlin and Ma, Kaixin and Yu, Wenhao and Dai, Yong and Zhang, Hongming and Lan, Zhenzhong and Yu, Dong},
  booktitle={Proceedings of the 62nd Annual Meeting of the Association for Computational Linguistics (Volume 1: Long Papers)},
  pages={6864--6890},
  year={2024}
}

@misc{claude_opus5,
    author={Anthropic},
    title={System Card: Claude Opus 5},
    year={2026},
    url={https://anthropic.com/claude-opus-5-system-card}
}

@misc{claude_sonnet5,
    author={Anthropic},
    title={System Card: Claude Sonnet 5},
    year={2026},
    url={https://anthropic.com/claude-sonnet-5-system-card}
}

@misc{google_gemini3_7_flash,
    author={Google},
    title={Introducing Gemini 3.7 Flash},
    year={2026},
    url={https://blog.google/innovation-and-ai/models-and-research/gemini-models/introducing-gemini-3-7-flash/}
}

@misc{google_gemini3_1_pro,
    author={Google},
    title={Gemini 3.1 Pro: A smarter model for your most complex tasks},
    year={2026},
    url={https://blog.google/innovation-and-ai/models-and-research/gemini-models/gemini-3-1-pro/}
}

@article{team2026kimi,
  title={Kimi k3: Open frontier intelligence},
  author={{Kimi Team} and Bai, Tongtong and Bai, Yifan and Bao, Yiping and Cai, Jianfeng and Cai, Xinyuan and Cao, Peizhou and Cao, Yuxuan and Chai, Ziwei and Charles, Y and others},
  journal={arXiv preprint arXiv:2607.24653},
  year={2026}
}

@misc{qwen3.8flashnext,
    title  = {{Qwen3.8-Flash-Next}: A New Architecture, Towards Ultimate Cost-Efficiency},
    author = {{Qwen Team}},
    month  = {August},
    year   = {2026},
    url    = {https://qwen.ai/blog?id=qwen3.8-flash-next}
}

@misc{qwen38,
    title = {{Qwen3.8-Max}: A New Bar for Coding and Cowork},
    url = {https://qwen.ai/blog?id=qwen3.8},
    author = {{Qwen Team}},
    month = {August},
    year = {2026}
}

@inproceedings{volkel2021eliciting,
  title={Eliciting and analysing users’ envisioned dialogues with perfect voice assistants},
  author={V{\"o}lkel, Sarah Theres and Buschek, Daniel and Eiband, Malin and Cowan, Benjamin R and Hussmann, Heinrich},
  booktitle={Proceedings of the 2021 CHI conference on human factors in computing systems},
  pages={1--15},
  year={2021}
}

@article{ye2025mobile,
  title={Mobile-agent-v3: Fundamental agents for gui automation},
  author={Ye, Jiabo and Zhang, Xi and Xu, Haiyang and Liu, Haowei and Wang, Junyang and Zhu, Zhaoqing and Zheng, Ziwei and Gao, Feiyu and Cao, Junjie and Lu, Zhengxi and others},
  journal={arXiv preprint arXiv:2508.15144},
  year={2025}
}

@article{chen2026appworld,
  title={AppWorld-UL: Benchmarking Diverse Agent-User Interactions for Tool-Use},
  author={Chen, Junzhi and Trivedi, Harsh and Pan, Jane and Zhang, Michael JQ and Srinivasan, Tejas and Balasubramanian, Niranjan and Sabharwal, Ashish},
  journal={arXiv preprint arXiv:2607.20536},
  year={2026}
}

@misc{openai2024embedding,
  author       = {OpenAI},
  title        = {New Embedding Models and {API} Updates},
  year         = {2024},
  month        = jan,
  url = {https://openai.com/index/new-embedding-models-and-api-updates/}
}

@article{kim2026persona2web,
  title={Persona2web: Benchmarking personalized web agents for contextual reasoning with user history},
  author={Kim, Serin and Lee, Sangam and Lee, Dongha},
  journal={arXiv preprint arXiv:2602.17003},
  year={2026}
}

@inproceedings{cai2025large,
  title={Large language models empowered personalized web agents},
  author={Cai, Hongru and Li, Yongqi and Wang, Wenjie and Zhu, Fengbin and Shen, Xiaoyu and Li, Wenjie and Chua, Tat-Seng},
  booktitle={Proceedings of the ACM on Web Conference 2025},
  pages={198--215},
  year={2025}
}

@article{qian2025userbench,
  title={Userbench: An interactive gym environment for user-centric agents},
  author={Qian, Cheng and Liu, Zuxin and Prabhakar, Akshara and Liu, Zhiwei and Zhang, Jianguo and Chen, Haolin and Ji, Heng and Yao, Weiran and Heinecke, Shelby and Savarese, Silvio and others},
  journal={arXiv preprint arXiv:2507.22034},
  year={2025}
}

@inproceedings{qian2024tell,
  title={Tell me more! towards implicit user intention understanding of language model driven agents},
  author={Qian, Cheng and He, Bingxiang and Zhuang, Zhong and Deng, Jia and Qin, Yujia and Cong, Xin and Zhang, Zhong and Zhou, Jie and Lin, Yankai and Liu, Zhiyuan and others},
  booktitle={Proceedings of the 62nd Annual Meeting of the Association for Computational Linguistics (Volume 1: Long Papers)},
  pages={1088--1113},
  year={2024}
}

\appendix

\clearpage

\counterwithin{figure}{section}  
\counterwithin{table}{section}   

\renewcommand{\thefigure}{\thesection.\arabic{figure}}
\renewcommand{\thetable}{\thesection.\arabic{table}}

\DoToC

\newpage

\section{Comparison with Existing Benchmarks}
\label{app:comparison}

We provide a comparison with existing benchmarks in \autoref{tab:related_work}.

\begin{table*}[h]
\centering
\small
\caption{Comparison of \benchmark with existing benchmarks.
\textbf{Inferable Info.}: tasks omit information that the agent can recover from the user's behavioral history (\pmark: user data exists in the environment but is not exposed to the agent as context).
\textbf{Ask-only Info.}: tasks omit information that can only be obtained by asking the user.
\textbf{Necessity-Aware Interaction Eval.}: the user-agent interaction is evaluated against what genuinely needs to be asked.  
}
\resizebox{1\textwidth}{!}{%
\begin{tabular}{llccc}
\toprule
\multirow{2}{*}{\textbf{Benchmark}} & \multirow{2}{*}{\textbf{Environment}} & \multirow{2}{*}{\textbf{Inferable Info.}} & \multirow{2}{*}{\textbf{Ask-only Info.}} & \textbf{Necessity-Aware} \\
&&&& \textbf{Interaction Eval.}\\
\midrule
\multicolumn{5}{l}{\textit{General computer-use and GUI agent benchmarks}} \\
Mind2Web~\citep{deng2023mind2web}            & Web            & \xmark & \xmark & \xmark \\
VisualWebArena~\citep{koh2024visualwebarena} & Web            & \xmark & \xmark & \xmark \\
OSWorld~\citep{osworld}                      & Desktop        & \xmark & \xmark & \xmark \\
AndroidWorld~\citep{androidworld}            & Mobile         & \xmark & \xmark & \xmark \\
MobileWorld~\citep{kong2026mobileworld}      & Mobile         & \xmark & \cmark & \xmark \\
\midrule
\multicolumn{5}{l}{\textit{Personalized agent benchmarks}} \\
PersonalWAB~\citep{cai2025large} & Web & \cmark & \xmark & \xmark \\
Persona2Web~\citep{kim2026persona2web} & Web & \cmark & \xmark & \xmark \\
PersonalAlign~\citep{lyu2026personalalign}   & Mobile         & \cmark & \xmark & \xmark \\
ME-Agent~\citep{wang2026me}                        & Mobile         & \cmark & \xmark & \xmark \\
Fingertip~\citep{yang2025fingertip}          & Mobile         & \cmark & \xmark & \xmark \\
iOSWorld~\citep{iosworld}                    & Mobile (iOS)   & \pmark & \xmark & \xmark \\
VitaBench 2.0~\citep{chen2026vitabench}     & Tool           & \cmark & \cmark & \xmark \\
KnowU-Bench~\citep{knowUBench}               & Mobile         & \cmark & \cmark & \xmark \\
\midrule
\multicolumn{5}{l}{\textit{Interactive agent benchmarks}} \\
IN3~\citep{qian2024tell} & Tool & \xmark & \cmark & \xmark \\
$\tau$-Bench~\citep{yao2024tau}              & Tool           & \xmark & \cmark & \xmark \\
$\tau^2$-Bench~\citep{barres2025tau2bench}   & Tool           & \xmark & \cmark & \xmark \\
UserBench~\citep{qian2025userbench} & Tool & \xmark & \cmark & \xmark \\
AmbigSWE~\citep{vijayvargiya2026ambig}    & Code           & \xmark & \cmark & \xmark \\
Ask-before-Plan~\citep{askbeforeplan}        & Tool           & \xmark & \cmark & \cmark \\
AppWorld-UL~\citep{chen2026appworld}     & Tool     & \xmark & \cmark & \cmark \\
\midrule
\textbf{\benchmark (ours)}                   & Mobile (iOS)   & \cmark & \cmark & \cmark \\
\bottomrule
\end{tabular}
}
\label{tab:related_work}
\end{table*}
\section{Environment Details}

\subsection{Mock Applications}
\label{app:mock_apps}

\benchmark builds on the executable iOS environment from iOSWorld~\citep{iosworld}, with \num{26} mock applications covering a wide range of everyday activities. These applications are populated with interconnected data associated with a coherent user persona. We verify and revise the seeded data to ensure they are consistent and can provide sufficient evidence for user preferences used in \benchmark.
\autoref{tab:apps} lists all applications by category.

\begin{table*}[h]
\centering
\small
\caption{The \num{26} mock applications in the \benchmark environment, grouped by category.}
\label{tab:apps}
\begin{tabular}{lcc}
\toprule
\textbf{Category} & \textbf{App} & \textbf{Description} \\
\midrule
\multirow{4}{*}{Food \& dining} & DineSpot & Restaurant search and table reservations \\
 & FreshCart & Grocery shopping and delivery \\
 & QuickBite & Food ordering and delivery from restaurants \\
 & TasteRank & Restaurant reviews and ratings \\
\midrule
\multirow{8}{*}{Work \& productivity} & LockedIn & Professional networking profile, connections, and job postings \\
 & Notes & Personal notes and lists \\
 & Weather & Weather forecasts \\
 & Clock & Alarms, timers, and world clocks \\
 & CloudSheets & Spreadsheet creation and editing \\
 & CloudDocs & Document creation and editing \\
 & CloudDrive & Cloud file storage and sharing \\
 & CloudSlides & Presentation creation and editing \\
\midrule
\multirow{3}{*}{Messaging \& email} & QuickChat & Personal messaging with contacts and group chats \\
 & TeamChat & Workplace messaging channels and direct messages \\
 & Mail & Email \\
\midrule
\multirow{3}{*}{Shopping \& payments} & MegaMart & Online retail shopping and order tracking \\
 & SplitPay & Peer-to-peer payments and shared expense splitting \\
 & MyBank & Bank accounts, transactions, and transfers \\
\midrule
\multirow{3}{*}{Travel \& transport} & CityRide & Ride-hailing with multiple ride tiers \\
 & SkyTrip & Flight search and booking, including cabin and seat selection \\
 & StayFinder & Lodging search and booking \\
\midrule
\multirow{3}{*}{Entertainment} & TicketBox & Tickets for concerts, sports, and other live events \\
 & Cinephile & Movie information, ratings, and watchlists \\
 & ScoreZone & Sports scores, schedules, and followed teams \\
\midrule
\multirow{2}{*}{Health \& fitness} & Trailblaze & Workout recording and activity logs (e.g., runs, rides, hikes) \\
 & CalTrack & Food logging and nutrition tracking \\
\bottomrule
\end{tabular}
\end{table*}

\subsection{Action Space}
\label{app:action_space}

Agents interact with the iOS simulator through a standard set of mobile-use actions for GUI interaction, along with an additional \texttt{ask\_user} action.
Table~\ref{tab:action_space} provides the complete action space and its parameter specifications. All screen coordinates are normalized to the range $[0,1000]$, with $(0,0)$ representing the top-left corner and $(1000,1000)$ the bottom-right corner. When an agent invokes \texttt{ask\_user}, its message is forwarded to the user simulator, and the simulator's response is returned to the agent as an observation.

\begin{table}[h]
\centering
\caption{Action space supported by iPrefBench. The first six actions provide standard mobile GUI interaction and task control, while \texttt{ask\_user} is introduced to enable interactive clarification with the simulated user.}
\label{tab:action_space}
\small
\begin{tabular}{@{}p{0.17\linewidth}p{0.29\linewidth}p{0.43\linewidth}@{}}
\toprule
\textbf{Action} & \textbf{Parameters} & \textbf{Description} \\
\midrule

\texttt{tap\_xy} &
\texttt{x}, \texttt{y}: integers in $[0,1000]$ &
Tap a specified screen position using normalized coordinates. \\

\texttt{type} &
\texttt{text}: string &
Enter text into the currently focused input field. The agent must focus the field before typing. \\

\texttt{swipe} &
\texttt{direction}: up, down, left, or right; optional \texttt{x}, \texttt{y} &
Swipe in the specified direction, optionally from a given starting position. \\

\texttt{home} &
None &
Return to the device's home screen. \\

\texttt{wait} &
\texttt{duration}: number (seconds; default: 2.0) &
Pause execution to allow content loading, animations, or other interface updates. \\

\texttt{stop} &
\texttt{answer}: string &
Terminate the task and return the agent's final response. \\

\midrule

\texttt{ask\_user} &
\texttt{question}: string &
Send a natural-language clarification message to the simulated user and receive its response. \\

\bottomrule
\end{tabular}
\end{table}

\subsection{Per-Task Environment Control}
\label{app:per_task_control}

All tasks in \benchmark share a single seeded environment and base activity history $\mathcal{H}$. However, some tasks, particularly the perturbation tasks (\S\ref{subsec:tasks}), require the environment to differ from this shared seed. \benchmark supports a small set of declarative per-task fields, which are applied at episode setup and reverted afterwards:
\begin{enumerate}
    \item \textbf{Data overrides} patch the data served by an app for the current task, e.g., marking an item out of stock, clearing a restaurant's availability, or extending an app's in-app history.
    \item \textbf{Log events} append synthetic activity events to the history $\mathcal{H}$ rendered for the current task only.
    \item \textbf{Profile overrides} update entries of the user profile $\mathcal{P}$ used by the simulator and the judges, for tasks where the user's present preference diverges from the historical pattern.
\end{enumerate}
For example, environment perturbations are implemented through data overrides, while preference perturbations require log events to change the agent-received user activity history, data overrides to make sure the on-screen app data is consistent, and profile overrides to reflect the updated preference for the user simulator and evaluators.
\section{User Details}

Following iOSWorld~\citep{iosworld}, all applications in the simulator are seeded with data around a single fictional persona, \textit{Jordan Avery}. \benchmark exposes information about this user to the agent in two ways: (1) a chronological user activity history $\mathcal{H}$, provided to the agent as prior context (Appendix~\ref{app:user_log}), and (2) an interactive user simulator that the agent can consult through \texttt{ask\_user} (Appendix~\ref{app:simulator}). The simulator is grounded in a structured user profile $\mathcal{P}$ derived from the seeded data (Appendix~\ref{app:user_profile}), which also serves as ground truth for evaluation but is never shown to the agent.

\subsection{User Activity History}
\label{app:user_log}

We construct the user activity history $\mathcal{H}$ by converting the seeded application data into a unified, chronological sequence of natural-language events.
Each event is represented by its timestamp followed by a natural-language description of the activity, which is converted from its metadata with an LLM (Claude Sonnet 5). Events are ordered chronologically, preserving temporal information that may be important for inferring recurring habits and evolving preferences.

The following are two example events from the activity history:

\begin{quote}
\small
[2025-08-01 16:55] in MyBank transferred \$300.00 from Total Checking to Savings for monthly savings.

[2026-02-19 02:06] in QuickChat received a message from Camille Hart in Brunch Crew: ``Alright, making this group official. Monthly brunch is non-negotiable.''
\end{quote}

By combining activities from different applications into a single history, $\mathcal{H}$ allows agents to infer user information beyond the context of any individual application. For example, payment transactions, restaurant orders, and restaurant ratings may jointly provide evidence of dining preferences, and messages and travel bookings may provide complementary information about travel plans.

The activity history $\mathcal{H}$ is provided to the agent as its only prior knowledge of the user. In the default evaluation setting, the agent receives the full history at the beginning of each task.
All inferable requirements in \benchmark are grounded in evidence available in $H$. In contrast, ask-only requirements cannot be resolved from the history and must be obtained through the \texttt{ask\_user} action.

\subsection{User Profile}
\label{app:user_profile}

We derive a structured user profile $\mathcal{P}$ for the persona, grounded in the seeded application data and the corresponding user activity history $\mathcal{H}$. The profile captures the user's identity, relationships, preferences, habits, and other personal information across \num{13} dimensions, comprising \num{92} entries in total.
Each entry records its value together with an evidence field that points to the supporting events in $\mathcal{H}$, and every piece of information in $\mathcal{P}$ is designed to be recoverable from the history.
Profile entries are first extracted with coding agents from the seeded data in $\mathcal{H}$, and manually verified against their evidence in $\mathcal{H}$ during the task-authoring process.
The structured profile serves as a hidden oracle for the user simulator and evaluation. It is not directly exposed to the agent under test.
Table~\ref{tab:user_profile} summarizes these domains and their representative contents.

\begin{table}[h]
\centering
\caption{The 13 dimensions of the structured user profile in \benchmark, with representative information covered by each dimension.}
\label{tab:user_profile}
\small
\setlength{\tabcolsep}{5pt}
\renewcommand{\arraystretch}{1.1}
\begin{tabular}{@{}p{0.34\linewidth}p{0.59\linewidth}@{}}
\toprule
\textbf{Dimension} & \textbf{Representative Information} \\
\midrule
Identity & Name, occupation, education, and skills. \\
Digital & Devices, system settings, and default applications. \\
Home and Household & Home and work locations, roommate, and household information. \\
People & Family, friends, colleagues, and social relationships. \\
Finances & Accounts, payment methods, recurring bills, and spending habits. \\
Health and Fitness & Nutrition goals, exercise habits, and fitness preferences. \\
Routines & Daily, weekly, and monthly routines and commitments. \\
Food and Dining & Cuisine preferences, favorite restaurants, recurring orders, and dietary requirements. \\
Entertainment and Culture & Movies, music, reading, concerts, and sports interests. \\
Hobbies & Hobby interests, collections, and related activities. \\
Travel and Transport & Flight, seating, lodging, and rideshare preferences. \\
Shopping & Preferred retailers, brands, apparel sizes, and delivery preferences. \\
Work and Career & Professional presence and career interests. \\
\bottomrule
\end{tabular}
\end{table}

\subsection{User Simulator}
\label{app:simulator}

We simulate the user with Claude Sonnet 5, prompted to role-play the persona (prompt in Appendix~\ref{app:prompts}).
The simulator receives three sources of information: (1) the user profile $\mathcal{P}$, with task-specific overrides applied when applicable; (2) the task-specific hidden context, i.e., the intended values of the task's ask-only requirements (e.g., travel dates or party size); and (3) optional simulator context, which supplements additional situational details about the current request to keep the simulator's replies coherent, without introducing additional evaluated requirements.

To ensure faithful evaluation of the agent's own ability to identify ambiguities, the simulator is instructed to answer only what the agent explicitly asks and never volunteer additional information. Human validation of these behaviors is reported in Appendix~\ref{app:user_sim_validation}. We present an example interaction between the agent and the user simulator below:

\begin{tcolorbox}[colback=gray!3, colframe=gray!50, title=Example Interaction between the User Simulator and the Agent, fonttitle=\bfseries\small, fontupper=\small]
\textbf{Task:} I went to a show last night and a friend covered both our tickets — send them my share. \\
\textbf{Task context (ask-only information):} Concert companion: Theo Nguyen; Share amount in USD: 85.

\medskip\hrule\medskip

\textbf{Agent} (\texttt{ask\_user}): Which friend covered both our show tickets last night? I couldn't find a ticket payment to you in the recent SplitPay feed, so I want to make sure I send your share to the right person.\\
\textbf{User:} Theo Nguyen.\\

\medskip
\textbf{Agent} (\texttt{ask\_user}): How much should I send Theo for your share of the tickets? I don't have the ticket price on hand, so tell me the amount (or the total and I'll send half).\\
\textbf{User:} \$85.
\end{tcolorbox}

\subsection{Human Validation of User Simulator}
\label{app:user_sim_validation}

We manually annotate all agent-user interactions produced by four representative agents in our experiments (Claude Sonnet 5, Gemini 3.7 Flash, Kimi K3, Qwen3.8-27B), yielding \num{89} question-answer pairs.
For each question-answer pair, we check it in two aspects:
\begin{itemize}
    \item \textbf{Grounding:} Whether every fact in the reply is supported by the user profile, hidden task context, or the reply honestly declines something the profile genuinely does not cover. \emph{Fabricated} (i.e., the reply states a concrete detail that appears nowhere in the profile or context), \emph{Contradicts} (i.e., the reply gives a value that conflicts with the profile or task context), \emph{Wrongly withheld} (i.e., the reply claims not to know, or refuses, when the answer is in fact available) and \emph{Can't tell} (i.e., the profile is ambiguous or the question is too vague to judge) are all counted as failures.
    \item \textbf{Scope:} Whether the reply answers everything asked and nothing more. \emph{Volunteered extra} (i.e., the reply adds information the agent did not request) and \emph{Incomplete} (i.e., the reply leaves part of the question unanswered) are all counted as failures.
\end{itemize}

\num{96.6}\% of pairs (86 of 89) are valid on both axes. Both grounding failures occur in tasks without ask-only requirements, where the simulator has no direct answer and must improvise for agent-produced unnecessary questions.
The single scope failure is a reservation task in which the agent asks three specific questions (date, party size, and restaurant) and the simulator answers all three but additionally supplies the time.

\section{Task Details}

\subsection{Task Collection}
\label{app:task_collection}



We construct \benchmark through a three-stage pipeline consisting of automated candidate generation, human refinement, and execution-based validation. The pipeline produces \num{188} initial candidate tasks, from which \tasknum tasks are retained in the final benchmark.

\paragraph{Stage 1: Automated Candidate Generation.}
We develop two task-authoring skills for coding agents to generate candidate tasks. The first skill generates personalized mobile-use tasks using the application source code, seeded user data, user activity history, structured user profile, and fully specified tasks from iOSWorld as references. Candidate tasks are designed to resemble realistic user requests while leaving user-dependent information underspecified.

The second skill focuses on generating tasks with three types of perturbations: instruction perturbations, preference perturbations, and environment perturbations. These tasks introduce cases in which an explicit instruction overrides historical behavior, recent evidence indicates a change in preference, or a preferred option becomes unavailable in the environment. Together, the two skills generate an initial pool of 188 candidate tasks.

\paragraph{Stage 2: Human Refinement.}
Eight human reviewers refine the generated candidates. They review the naturalness of each user instruction, verify that the requested operations are feasible in the environment, and identify the user-dependent information omitted from the instruction.

For each omitted requirement, reviewers determine its resolution source and ensure that it is well grounded. An inferable requirement must be supported by a dominant pattern or an explicit fact in $\mathcal{H}$; when the supporting evidence is weak or mixed, reviewers revise the seeded data so that $\mathcal{H}$ supports a clear preference. An ask-only requirement must not be recoverable from $\mathcal{H}$, and its intended value is specified in the task-specific hidden context. Whenever possible, this value is chosen to differ from the app's pre-filled default, so that silently accepting the default or guessing is unlikely to succeed by chance.  Reviewers revise ambiguous requirements and the corresponding task rubrics to ensure that each requirement has a well-defined ground-truth value and resolution source. For tasks whose conditions must diverge from the shared environment, such as perturbation tasks, reviewers also specify or revise the corresponding per-task overrides (Appendix~\ref{app:per_task_control}).

The reviewers also remove duplicate tasks and refine the candidate pool to improve coverage of different applications, user preferences, and task scenarios.

\paragraph{Stage 3: Execution-Based Validation.}
Two experienced human validators execute the refined tasks in the
iOS simulator. This stage checks that the requested operations
can be completed in the environment and that the task rubrics
correctly specify the expected outcomes. Remaining issues in
task instructions, environmental feasibility, and rubrics are
corrected during validation.

Following refinement and validation, we retain \tasknum tasks in
the final benchmark. Each retained task undergoes both human
refinement and execution-based validation.

\subsection{Benchmark statistics}
\label{app:statistic}

\autoref{tab:stats} summarizes the statistics of \benchmark, including the user environment, task composition, and evaluation rubrics. A task involves \emph{cross-app preference transfer} if the evidence for at least one of its inferable requirements comes from an app other than the one in which the requirement must be applied (e.g., inferring a dining preference from restaurant ratings in TasteRank and applying it to a reservation in DineSpot).

\begin{table}[h]
  \centering
  \small
  \caption{Statistics of \benchmark.}
  \label{tab:stats}
  \begin{tabular}{lr}
  \toprule
  \textbf{Statistic} & \textbf{Value}\\
  \midrule
  \multicolumn{2}{l}{\textit{User persona}} \\
  \quad Activity events    & \eventnum   \\
  \quad Mock apps with seeded user data & 26 \\
  \quad Profile dimensions / entries   & \num{13} / \num{92}   \\
  \midrule
  \multicolumn{2}{l}{\textit{Tasks}} \\
  \quad Tasks                                 & \tasknum     \\
  \quad \quad Tasks with cross-app preferences & \num{35} (\num{26.9}\%) \\
  \quad \quad Tasks with $\geq$1 ask-only slot & \num{76} (\num{58.5}\%) \\
  \quad \quad Perturbed: inst. / pref. / env. & 11 / 10 / 13 \\
  \quad Apps covered & \num{26} \\
  \quad Profile entries covered & \num{67} (\num{72.8}\%) \\
  \midrule
  \multicolumn{2}{l}{\textit{Rubrics}} \\
  \quad Avg. \# Rubric criteria                & \num{4.96} \\
  \quad \quad action-completion               & \num{2.09} \\
  \quad \quad personalization (inferable / ask-only)                 & \num{2.87} (\num{1.98} / \num{0.89}) \\
  \bottomrule
  \end{tabular}
  \end{table}

\begin{figure}
    \centering
    \includegraphics[width=0.8\linewidth]{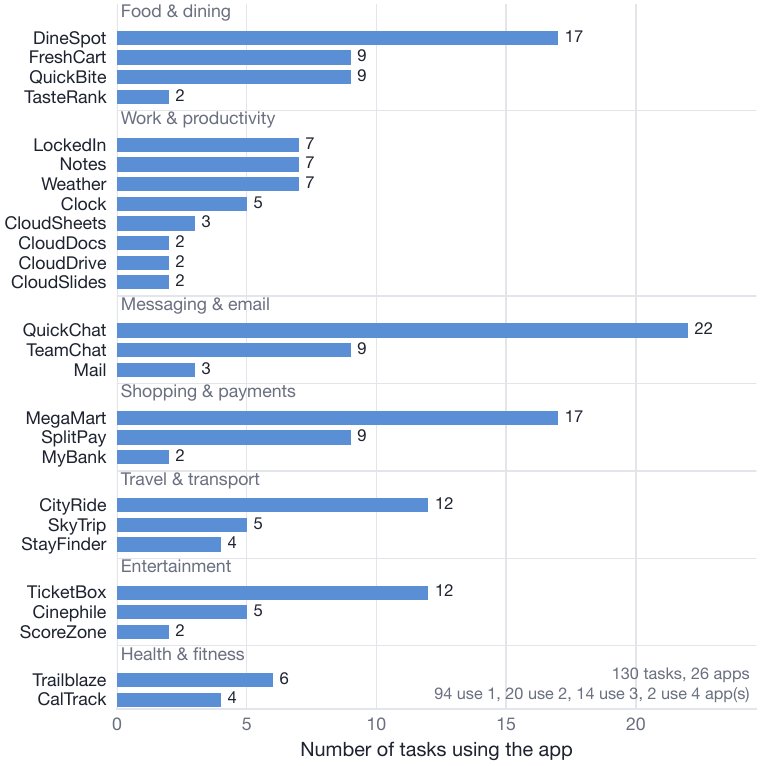}
    \caption{Distribution of apps used by \benchmark tasks. Bars count the tasks that require the app, grouped by app category and sorted within each category. Some tasks span multiple apps.}
    \label{fig:task_app_distribution}
\end{figure}

The tasks cover all 26 applications in the \benchmark environment. \autoref{fig:task_app_distribution} shows the number of tasks involving each application, grouped by functional category.

\section{Rubric and Interaction Judge Details}
\label{app:judge}

\subsection{Task-specific Rubrics}

Each task is associated with a rubric containing two types of
criteria: action-completion and personalization criteria.

Action-completion criteria specify the operations required to
complete the task, independently of the values selected for
user-dependent parameters. Personalization criteria assess whether
the agent makes the correct user-dependent decisions. Each
personalization criterion is additionally annotated with its
information resolution source: \emph{inferable}, if the intended
value can be recovered from the user activity history $\mathcal{H}$, or
\emph{ask-only}, if the value is unavailable in $\mathcal{H}$ and must
be obtained through user interaction.

\subsection{Rubric-Based Completion Judge}

We use an LLM-based judge (with Claude Sonnet 5) to evaluate each rubric criterion independently. The judge receives the task instruction, the complete execution trajectory, and the agent's final answer.
The trajectory contains the executed actions and screenshots at each step, together with a screenshot of the final interface state. The judge additionally receives the ground-truth user profile and task-specific hidden context when applicable.
The judge is also given the run time of the task, i.e., the simulator clock during execution, to resolve relative temporal expressions such as ``tomorrow'' or ``next Saturday.''

For each criterion, the judge returns a binary label and a supporting explanation. A criterion passes only when the trajectory provides clear evidence that it has been satisfied.

We compute action-completion and personalization scores as the fractions of satisfied criteria within their respective categories. Both action-completion and personalization scores are macro-averaged across tasks. Personalization is further reported separately for inferable and ask-only criteria, which are macro-aggregated over tasks containing at least one criterion of the corresponding type. A task is considered fully successful only if all of its action-completion and personalization criteria are satisfied.

\subsection{Interaction Judge}
To evaluate whether agents ask appropriate clarification questions, we analyze the \texttt{ask\_user} messages produced during each task.
The interaction judge receives all clarification questions issued during a task and the task's annotated set of ask-only information. It identifies the individual pieces of information requested by the agent and semantically matches each piece to an ask-only requirement, if applicable.

One information piece corresponds to a single decision or value the user must supply. A question requesting several independent values is decomposed into multiple pieces. Conversely, a question offering several candidate answers for the same decision is counted as a single piece. Repeated requests are counted separately, even when they concern the same information.

For each extracted piece, the judge returns a short description and either the matched ask-only requirement or a null match. Matching is semantic rather than based on exact wording. Requests for information that is inferable from history, irrelevant to the task, or otherwise not among the ask-only requirements receive null matches.

Both interaction Recall and Precision are micro-aggregated: we sum the number of distinct ask-only requirements elicited, the total number of ask-only requirements, and the total number of information requests across all tasks.

\paragraph{Format-Level Message Quality.}
We also examine the format-level quality of each agent message along three criteria: \emph{specificity} (i.e., it is clear which piece of information is being requested)~\citep{vijayvargiya2026ambig}, \emph{answerability} (i.e., the user can answer from knowledge they plausibly have, given the context the message supplies)~\citep{vijayvargiya2026ambig}, and \emph{option provision} (i.e., when the answer space is a small closed set, the message surfaces the candidates rather than leaving the user to enumerate them)~\citep{peng2025morae}. An LLM judge scores each criterion on every message issued by the agent whenever the criterion applies, and the overall quality score is the message-level average over the applicable criteria.

\begin{table*}[h]
\centering
\caption{Message quality evaluation, which is the message-level average over the three criteria. Each criterion is scored only on the messages where it applies (Option Provision is only applicable to questions with a small closed set of answers).}
\small
\label{tab:interaction_quality}
\resizebox{1\textwidth}{!}{%
\begin{tabular}{lccccc}
\toprule
\textbf{Model} & \textbf{Total \#Ask} & \textbf{Overall Quality} & \textbf{Specificity} & \textbf{Answerability} & \textbf{Option Provision} \\
\midrule
Claude Opus 5      & \num{46} & \num{98.55}  & \num{100.00} & \num{100.00} & \num{93.75}  \\
Claude Sonnet 5    & \num{31} & \num{100.00} & \num{100.00} & \num{100.00} & \num{100.00} \\
Gemini 3.1 Pro     & \num{26} & \num{100.00} & \num{100.00} & \num{100.00} & \num{100.00} \\
Gemini 3.7 Flash   & \num{11} & \num{93.94}  & \num{100.00} & \num{100.00} & \num{71.43}  \\
Kimi K3            & \num{20} & \num{100.00} & \num{100.00} & \num{100.00} & \num{100.00} \\
Qwen3.8-27B        & \num{27} & \num{100.00} & \num{100.00} & \num{100.00} & \num{100.00} \\
Qwen3.8-Flash-Next & \num{34} & \num{98.53}  & \num{100.00} & \num{97.06}  & \num{100.00} \\
\bottomrule
\end{tabular}
}
\end{table*}

As shown in~\autoref{tab:interaction_quality}, all evaluated agents achieve near-perfect format quality. This indicates that the bottleneck in user interaction is not \emph{how} agents ask but \emph{whether} they ask: agents phrase their questions well once they decide to ask, yet, as shown in \S\ref{sec:exp}, they frequently fail to recognize when a question is necessary.
Among the three criteria, option provision is the only one that poses a modest challenge. Agents occasionally pose questions without listing possible options (e.g., in-stock colors of an item, the seating sections available on a venue map), although the options are visible on the screen, shifting the burden of inspecting options back to the user.

\subsection{Validation of LLM-as-a-Judge}
\label{app:human_validation}

\paragraph{Human Validation.} To assess the reliability of our LLM-based evaluation, we compare judge outputs against human annotations on 100 sampled agent trajectories. The trajectories are divided into two disjoint subsets of 50, each assigned to one human annotator. Each annotator evaluates the satisfaction of individual rubric criteria and the information requested in agent-user interactions.

For personalized task completion, we measure agreement at both the rubric-criterion level and the task-success level. At the criterion level, agreement between human annotations and the LLM judge reaches 95.7\%, with Cohen's $\kappa=0.89$. At the task level, agreement reaches 97.0\%, with $\kappa=0.94$.

For interaction Recall, we compare the required-to-ask slots identified as elicited by the interaction judge with human annotations. Agreement reaches 95.2\% ($\kappa=0.90$).
For interaction Precision, direct comparison at the level of individual extracted information pieces is not well-defined because the number and description of extracted pieces may differ between human annotators and the LLM judge. Instead, we compare the number of extra information requests identified by each, including requests for inferable or irrelevant information and redundant requests. Agreement on extra-request counts reaches 93.9\% ($\kappa=0.82$). Since these counts are ordinal, we additionally compute weighted Cohen's $\kappa$, which accounts for the magnitude of disagreements between counts and reaches \num{0.91}.

These results support the reliability of our LLM-based evaluation for both personalized task completion and information-request behavior.

\paragraph{Cross-Judge Validation.}
We further re-evaluate all trajectories in~\autoref{tab:main} with Gemini 3.7 Flash as an alternative judge.
For personalized task completion, the two judges agree on \num{95.9}\% of rubric criteria ($\kappa=0.91$) and \num{96.4}\% of task-level success labels ($\kappa=0.92$), with similar agreement on action-completion (\num{95.7}\%, $\kappa=0.87$) and personalization criteria (\num{95.9}\%, $\kappa=0.92$).
For interaction evaluation, they agree on \num{99.1}\% of ask-only requirements for recall ($\kappa=0.98$) and on \num{98.6}\% of per-task extra-request counts for precision ($\kappa=0.90$; weighted $\kappa=0.87$).
Overall, the results remain consistent with an alternative judge, demonstrating the consistency and reliability of our LLM-as-a-Judge.
\section{Experiment Details}
\label{app:exp}

\subsection{Evaluated Agents}
\label{app:agents}

We evaluate seven frontier agents in our experiments: Claude Opus 5~\citep{claude_opus5}, Claude Sonnet 5~\citep{claude_sonnet5}, Gemini 3.1 Pro~\citep{google_gemini3_1_pro}, Gemini 3.7 Flash~\citep{google_gemini3_7_flash}, Kimi K3~\citep{team2026kimi}, Qwen3.8-27B~\citep{qwen38}, and Qwen3.8-Flash-Next~\citep{qwen3.8flashnext}.

We additionally attempted to evaluate open-weight specialized GUI agents under our standard full-history setting. Some (e.g., UI-TARS-1.5-7B~\citep{qin2025ui} and GUI-Owl-1.5-32B-Instruct~\citep{ye2025mobile}) could not accommodate the complete user activity log within their supported context windows.
Some (e.g., MAI-UI-8B~\citep{zhou2025maiUI}) could accommodate the input length but frequently produced malformed or unparsable actions when the full history was included, resulting in a low Action Completion score of \num{16.9}\%. Since these failures occur at the basic action-execution level, the resulting personalization scores would be heavily confounded by execution failures rather than reflecting the agents' ability to infer preferences or seek clarification. We therefore exclude these models from the main comparative evaluation and focus on agents that can reliably operate under the same full-history setting.

\subsection{Error Analysis}
\label{app:error_analysis}

To understand how personalization fails, we manually label the error types in every failed trajectory of Claude Sonnet 5 (79 failed tasks) and Qwen3.8-27B
(96 failed tasks), which are representative of closed and open agents.
We first categorize personalization-related errors based on two axes: whether the underspecified information is ask-only or inferable, and whether the agent recognized the slot as open at all or resolved it incorrectly on its own without the user.
This yields four error types:
\begin{itemize}
    \item \textbf{Silent Default} (e.g.,~\autoref{fig:silent_default}): an ask-only requirement is never considered, and the app's pre-filled value is accepted directly as fact;
    \item \textbf{Missed Inference} (e.g.,~\autoref{fig:missed_inference}): an inferable preference that the task depends on is never considered;
    \item \textbf{Silent Guess} (e.g.,~\autoref{fig:silent_guess}): the agent does treat an ask-only detail as open but fills it in incorrectly on its own, by guessing at the current situation and user intention, e.g., based on history, plausibility, instead of asking;
    \item \textbf{Wrong Inference} (e.g.,~\autoref{fig:wrong_inference}): the agent reasons about a preference based on $\mathcal{H}$ but resolves it to the wrong value. 
\end{itemize}
\autoref{fig:error_analysis} shows the error distributions.
Note that a single trajectory may exhibit multiple error types.

The results show that failures on \benchmark are mostly personalization failures rather than navigation failures: \num{89}\% of Sonnet 5's and \num{84}\% of Qwen3.8-27B's failed trajectories contain at least one of the four personalization errors, consistent with the gap between action completion and personalization in~\autoref{tab:main}.
Although Qwen3.8-27B fails on more tasks, the error distributions of these two agents are close, suggesting that these are shared behaviors of agents optimized for autonomous execution rather than a weakness of any
single model.

\subsection{Additional Case Studies}
\label{app:add_case_studies}

In environment-perturbation tasks, the expected behavior is to recognize that the preferred option is unavailable, surface this to the user, and then either proceed with an acceptable alternative or stop when no substitute is acceptable (\S\ref{subsec:tasks}). We observe three different failure patterns of handling such unavailability:
(1) after finding the item unavailable, the agent keeps attempting other ways to obtain the same item (e.g.,~\autoref{fig:keep_trying});
(2) the agent directly substitutes an alternative without consulting the user (e.g.,~\autoref{fig:silent_substition});
and (3) the agent skips the unavailable item, without considering alternatives or consulting the user (e.g.,~\autoref{fig:abandon}).

\section{Prompts}
\label{app:prompts}

\subsection{User Simulator}

\begin{tcolorbox}[colframe=mygreen!50!black, colback=mygreen!10!white, title=Prompt for User Simulator, breakable]
\small

You are \{persona\}, the owner of this phone. An assistant is performing a task for you on the phone and may ask you questions to clarify or confirm. Answer naturally and consistently with your profile.

\medskip

Ground every answer in the profile below and what this person would plausibly know. Do NOT invent or fabricate details that are not supported by the profile. If a question is not covered by the profile or you would not know the answer, say so honestly. If a question is too vague or broad for you to tell what is actually being asked, say that it's unclear and ask what they mean rather than guessing or offering all related information. Do not perform the task yourself, and never reveal that you are an AI.

\medskip

Reply with EXACTLY one JSON object and no other text: \{``reply'': ``what you say''\}

\medskip

\textbf{Profile:}
\begin{quote}
    \{user\_profile\}
\end{quote}

\medskip

\textbf{You asked the assistant to do the following for you:}

\begin{quote}
\{user\_instruction\}
\end{quote}

The assistant's questions will be about this task. Answer in that context, but still only answer what's asked — don't volunteer details the assistant hasn't asked for.

\medskip

\textbf{For THIS specific request, the concrete details you have in mind (these are NOT in the profile above) are:}
\begin{quote}
    \{task\_context\}
\end{quote}

Reveal each of these only if the assistant explicitly asks for it; never volunteer them.

\medskip

\textbf{The current situation for this request:}

\begin{quote}
    \{simulator\_context\}
\end{quote}
Answer consistently with this. Say only what's asked and stay in character; don't volunteer extra details.

\end{tcolorbox}

\subsection{Personalized Mobile-Use Agent}

\begin{tcolorbox}[colframe=myblue!50!black, colback=myblue!10!white, title=Prompt for Agent, breakable]
\small

You are an iOS Simulator control agent. Reply with exactly one JSON object and no other text.

\medskip

Schema: \{``summary'':``your reasoning for this turn'',``actions'':[\dots]\}

\medskip

Rules:
\begin{itemize}
    \item You are in an agentic loop. After your actions execute, you will receive a new observation.
    \item Use the normalized 0-1000 coordinate space for all coordinates, where (0,0) is the top-left and (1000,1000) is the bottom-right.
    \item Do not chain many actions unless the next state is highly predictable. Only use stop when the task is fully complete.
    \item If uncertain about the UI — where an element is, what a control does — make the best grounded attempt and use the next observation to recover.
\end{itemize}

You are acting on behalf of a specific user, and you are given their activity log for context. Their requests may be underspecified, and you should act the way they would. When a request leaves a detail open, first resolve it from what you know about the user rather than choosing an arbitrary default. When a detail genuinely matters and you cannot settle it confidently from what you know about them, use ask\_user to ask or confirm instead of guessing.

\medskip

You are in vision-only mode: you do not have access to accessibility identifiers, UI metadata, or hidden state. Carefully estimate the centre of the element you want to tap in 0-1000 coordinates.

\medskip

Action schema — emit a JSON object with keys ``actions'' (array) and ``summary'' (string).

Each action is a JSON object with ``type" and the parameters listed below.
Use ONLY the action "type" values listed below — never invent other types
(there is no double\_tap/triple\_tap/hover/etc.). To select-all-then-replace
text, tap the field and use "type" with the new text.

\medskip

\textbf{Available actions:}

\medskip
\quad  \texttt{tap\_xy}
\begin{itemize}
    \item Tap at a screen position using coordinates in 0-1000 space (0=top/left edge, 1000=bottom/right edge).
    \item Parameters: x (integer, 0-1000), y (integer, 0-1000)
    \item Example: \{``type'': ``tap\_xy'', ``x":500, ``y":200\}
\end{itemize}

\quad \texttt{type}
\begin{itemize}
    \item Type text into the currently focused input field. You must tap a text field first to focus it. IMPORTANT: Ensure the text field is focused before typing. If typing produces no result or duplicate text, tap the field again and retry.
    \item Parameters: text (string, the text to enter)
    \item Example: \{``type'': ``type'', ``text'': ``hello world''\}
\end{itemize}

\quad  \texttt{swipe}
\begin{itemize}
    \item Swipe in a direction, optionally from a specific origin point.
    \item Parameters: direction (string, one of ``up''/``down''/ ``left''/``right''), x (integer, optional origin 0-1000), y (integer, optional origin 0-1000)
    \item Example: \{``type'': ``swipe'', ``direction'': ``up''\}
    \item Example with origin: \{``type'': ``swipe'', ``direction'': ``left'', ``x":800, ``y":500\}
\end{itemize}

\quad  \texttt{home}
\begin{itemize}
    \item Press the home button to return to the home screen.
    \item Parameters: none
    \item Example: \{``type'': ``home''\}
\end{itemize}

\quad  \texttt{wait}
\begin{itemize}
    \item Pause for a specified duration (seconds). Use this when you need to wait for content to load, animations to complete, or a delay before the next action.
    \item Parameters: duration (number, seconds to wait; default 2.0)
    \item Example: \{``type'': ``wait'', ``duration'':3.0\}
\end{itemize}

\quad  \texttt{stop}
\begin{itemize}
    \item End the task and provide your final answer. ONLY use this when you have fully completed the goal. Do NOT stop early — keep trying alternative approaches if your current actions are not working.
    \item Parameters: answer (string, REQUIRED — if the task asks you to return, tell, or report any information, you MUST include all requested data here. This is how you communicate results back to the user.)
    \item Example: \{``type'': ``stop'', ``answer'': ``The movie's year is 1972 and its rating is 9.2/10. It has been added to the Wishlist.''\}
\end{itemize}

\quad  \texttt{ask\_user}
\begin{itemize}
    \item Ask the user a question to get clarification or confirmation. Use this when a choice or the task's requirements are unclear, or when you don't have enough information to complete the task confidently.
    \item Parameters: question (string)
    \item Example: \{``type'': ``ask\_user'', ``question'': ``Which card should I pay with — the Visa or the Amex?''\}
\end{itemize}

\medskip

\textbf{Output requirements:}

\begin{itemize}
    \item Return valid JSON only. Do not use markdown or code fences.
    \item Do NOT use $<$function\_calls$>$, $<$invoke$>$, or any XML-style tool tags — respond with plain JSON.
    \item ALWAYS include a non-empty "summary" BEFORE the actions: concisely state your reasoning for this turn — what you observe, what you infer (including from the user context above, if any), and why you chose these actions. Keep it concise. Never return actions without a "summary".
\end{itemize}

\medskip

\textbf{User Activity Log:}

\begin{quote}
    \{user\_activity\_log\}
\end{quote}

\medskip

\textbf{Goal:}
\begin{quote}
    \{goal\}
\end{quote}
\end{tcolorbox}

\subsection{Rubric-based Judge}

\begin{tcolorbox}[colframe=red!50!black, colback=red!10!white, title=Prompt for Rubric-based Judge, breakable]
\small

You are a strict evaluation judge for an iOS agent benchmark. Read the goal, the
trajectory, and any ground-truth (the user profile and request specifics) in the
user message, then reply with EXACTLY ONE JSON object in the format the user
message specifies. Do NOT use any agent action schema; do NOT wrap your answer in
`actions' or `summary'. Return only the judge JSON object.

\medskip

\textbf{Goal:}
\begin{quote}
    \{goal\}
\end{quote}

\medskip

The task was run at \{run\_time\}. This is the phone's clock during the run — resolve any relative date in the goal or the rubric criteria (for example ``next Saturday'', ``the upcoming weekend'', ``tomorrow'') against it.

\medskip

You are evaluating whether an iOS agent successfully completed the above goal. The agent executed \{N\} steps. Full trajectory with per-step screenshots:

\begin{quote}
    \{trajectory\}
\end{quote}

\medskip

The agent's final answer was: ``\{agent\_answer\}''

\medskip

\textbf{The user's profile (GROUND TRUTH for preference/personalization criteria).}

Judge any criterion about the user's preferences strictly for consistency with THIS profile — not your own assumptions:

  \begin{quote}
      \{user\_profile\}
  \end{quote}

\medskip

\textbf{This request's specifics (GROUND TRUTH for intent criteria such as party
  size, day, or time — these are what the user actually wanted for this request):}
\begin{quote}
    \{task\_specific\_hidden\_context\}
\end{quote}

\medskip

Evaluate each of the following rubric criteria individually:
\begin{enumerate}
    \item \{criterion\_1\}
    \item \{criterion\_2\}
    \item \dots
\end{enumerate}

\medskip
  Respond with ONLY a JSON object (no code fences) in this format:

  \begin{quote}
      \{
    
    \quad ``success'': true/false,
    
    \quad ``reasoning'': "overall assessment of whether the task was completed",
    
    \quad ``rubric\_results'': [
    
    \quad \quad  \{``criterion'': ``criterion\_text'', ``satisfied'': true/false, ``reasoning'': ``why''\},
    
    \quad \quad \dots
    
    \quad ]
  
  \}
  \end{quote}
  
  success=true means the goal is fully and completely achieved. For each rubric
  criterion, set satisfied=true only if there is clear evidence in the trajectory
  and screenshots that the criterion was met. Treat each rubric criterion as the
  AUTHORITATIVE and INDEPENDENT definition of success for that item: judge it solely
  by its own literal wording, against the evidence and the ground-truth
  profile/specifics. Do NOT make a criterion stricter or looser based on how the
  goal is phrased — the goal is only what the agent was asked; the criterion is what
  counts. If the goal's wording seems to demand more (or less) than a criterion
  states, follow the criterion. When a criterion refers to the user's preferences or
  to this request's specifics, judge it strictly against the user profile and request
  specifics provided above — not your own assumptions. For tasks that ask a question,
  evaluate whether the agent's final answer is correct.
  
\end{tcolorbox}

\subsection{Interaction Judge}

\begin{tcolorbox}[colframe=yellow!50!black, colback=yellow!10!white, title=Prompt for Interaction Judge, breakable]
\small

You are a strict evaluation judge for the interaction quality of an iOS agent's
messages to the user. Reply with EXACTLY ONE JSON object in the required format.

\medskip

An iOS agent asked the user these clarifying questions:
\begin{itemize}
    \item \{question\_1\}
    \item \{question\_2\}
    \item \dots
\end{itemize}

The user's request had these REQUIRED-TO-ASK information slots (the agent could not know them without asking):

\begin{quote}
    \{task\_specific\_hidden\_context\}
\end{quote}

First, enumerate the pieces of information the agent requested across all the
  questions above. A piece is ONE thing the user has to decide or supply — one
  decision, or one value. Split a question that bundles several independent pieces
  (things the user would have to answer separately) into separate pieces. Candidate
  answers enumerated for ONE decision are ONE piece, not several: a question offering
  the user a choice between alternatives, or listing options to pick from, requests a
  single piece however many options it lists. Count every ask: if the same piece is
  requested more than once (e.g. re-asked in a later message), list it each time —
  do NOT merge repeats.

  \medskip
  
  For EACH piece, decide which required-to-ask slot (if any) it requests (semantic
  match, not exact wording). Use the EXACT slot name from the list above, or null if
  the piece does not correspond to any required slot.
  
  \medskip
  Respond with ONLY this JSON (no code fences): \{``info\_pieces'': [\{``piece'': ``short
  description'', ``matched\_slot'': ``slot name or null''\}]\}

\end{tcolorbox}

\subsection{Pre-execution Clarification}

\begin{tcolorbox}[colframe=mydarkgray!50!black, colback=mydarkgray!10!white, title=Prompt for Pre-execution Clarification, breakable]
\small

You are an assistant about to perform a task on a user's phone on their behalf.
You are given the user's activity log (context about their habits and preferences) and the task.

\medskip

  BEFORE starting, identify the choices or parameters this task leaves
  UNDERSPECIFIED and that depend on the USER'S PREFERENCES or intent — NOT
  objective/mechanical details you can figure out yourself.

\medskip

  First try to resolve each open detail from the user's activity log. Only ask about
  details that genuinely matter and that you CANNOT settle confidently from the log
  or from what you know about the user.

  \medskip
  
  Write ONE short, natural clarifying question for each such detail.

\medskip

Ask about every genuinely underspecified detail — but only those you cannot
    settle from the log; do not pad with unnecessary questions. If nothing needs
    clarifying, return an empty list.

\medskip

  Reply with EXACTLY one JSON object and no other text: \{``clarifying\_questions":
  [``...'', ``...'']\}

  \medskip

  \textbf{User Activity Log:}

\begin{quote}
    \{user\_activity\_log\}
\end{quote}

  \textbf{Task goal:}
  
  \begin{quote}
      \{goal\}
  \end{quote}

\end{tcolorbox}

\clearpage   
\begingroup
\renewcommand{\thefigure}{F.\arabic{figure}}
\renewcommand{\theHfigure}{errcase.\arabic{figure}}  
\setcounter{figure}{0}

\begin{figure*}
    \centering
    \includegraphics[width=0.9\linewidth]{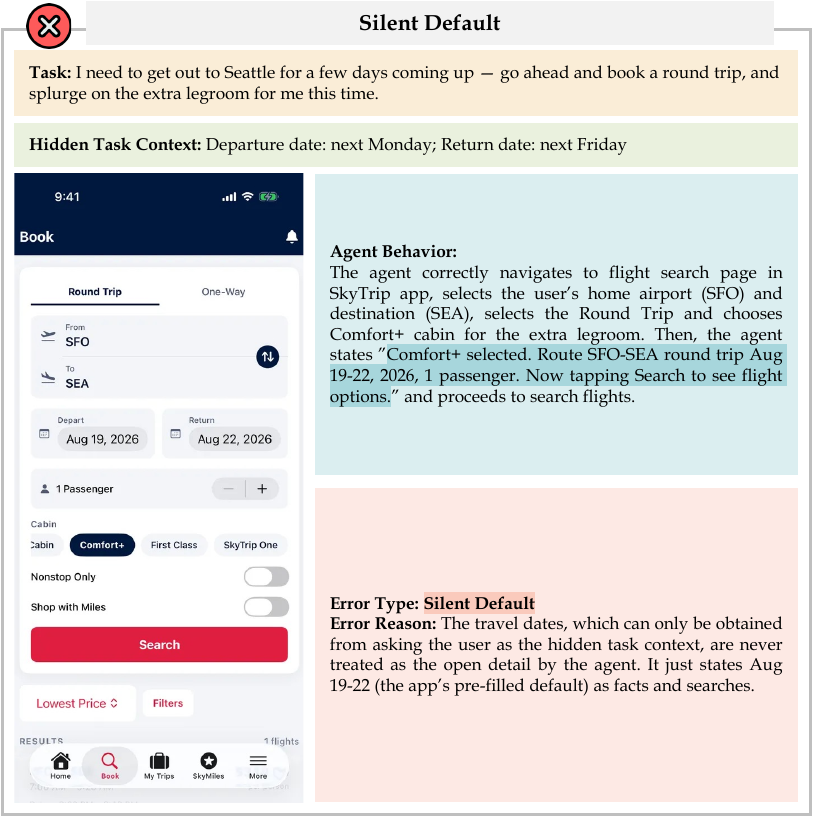}
    \caption{An example of \textit{Silent Default}. When asked to book a round trip for ``a few days coming up'' without specific dates, the agent directly uses the app's default departure and return dates as if they were the user's, never treating travel dates as an open detail that needs to be resolved.}
    \label{fig:silent_default}
\end{figure*}

\begin{figure*}
    \centering
    \includegraphics[width=0.9\linewidth]{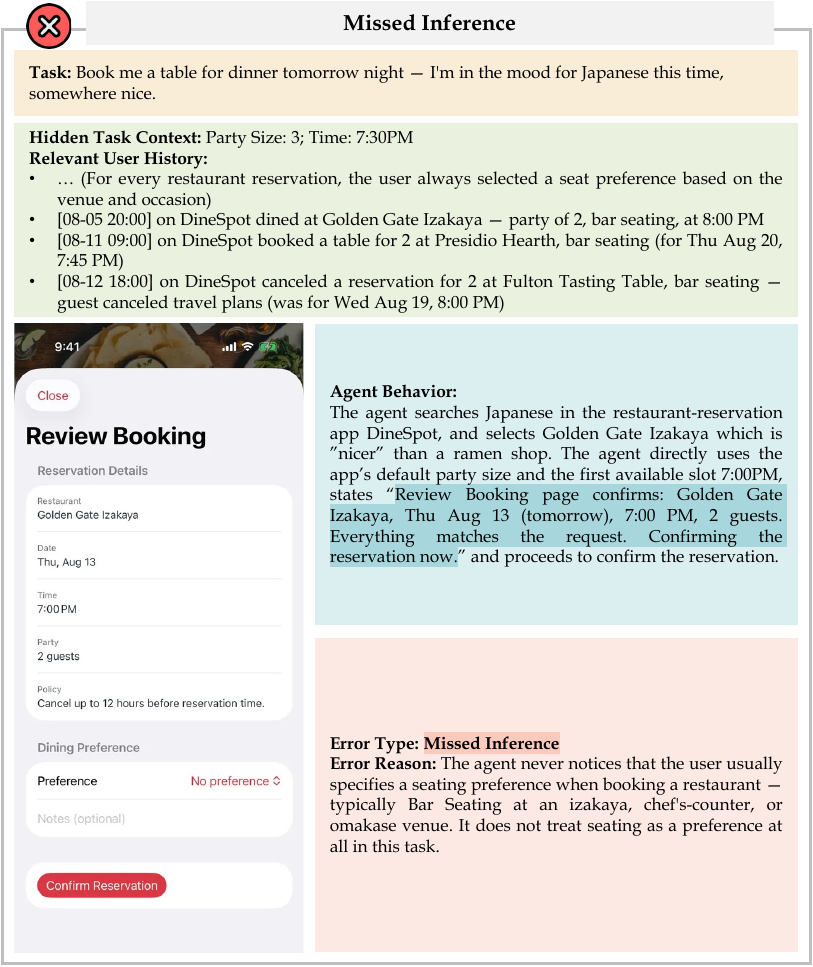}
    \caption{An example of \textit{Missed Inference}. While the user sets a seat preference on every restaurant booking, and selects Bar Seating for an izakaya, chef's-counter or omakase venue, the agent never notices this user preference, so seating is absent from its reasoning entirely. When booking at an izakaya, the agent leaves the reservation with ``No Preference''.}
    \label{fig:missed_inference}
\end{figure*}

\begin{figure*}
    \centering
    \includegraphics[width=0.9\linewidth]{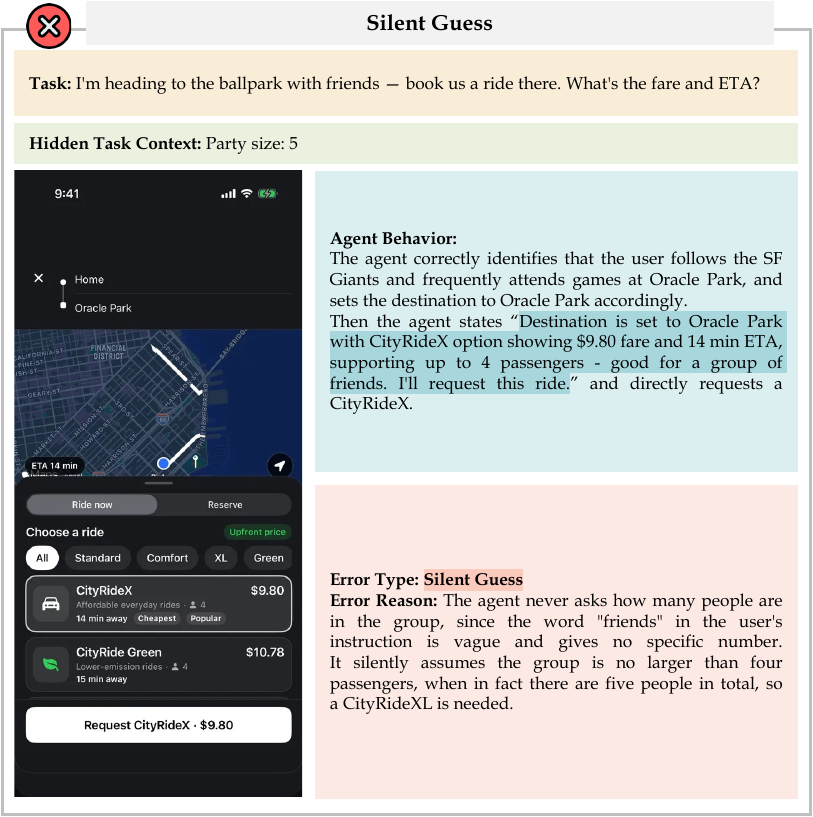}
    \caption{An example of \textit{Silent Guess}. The user instruction explicitly mentions that the ride is ``with friends'' without the specific group size, and only the user knows the actual group size (i.e., \num{5}) in the current situation. However, the agent directly assumes ``with friends'' as a group of at most four and books a ride that seats one person too few.}
    \label{fig:silent_guess}
\end{figure*}

\begin{figure*}
    \centering
    \includegraphics[width=0.9\linewidth]{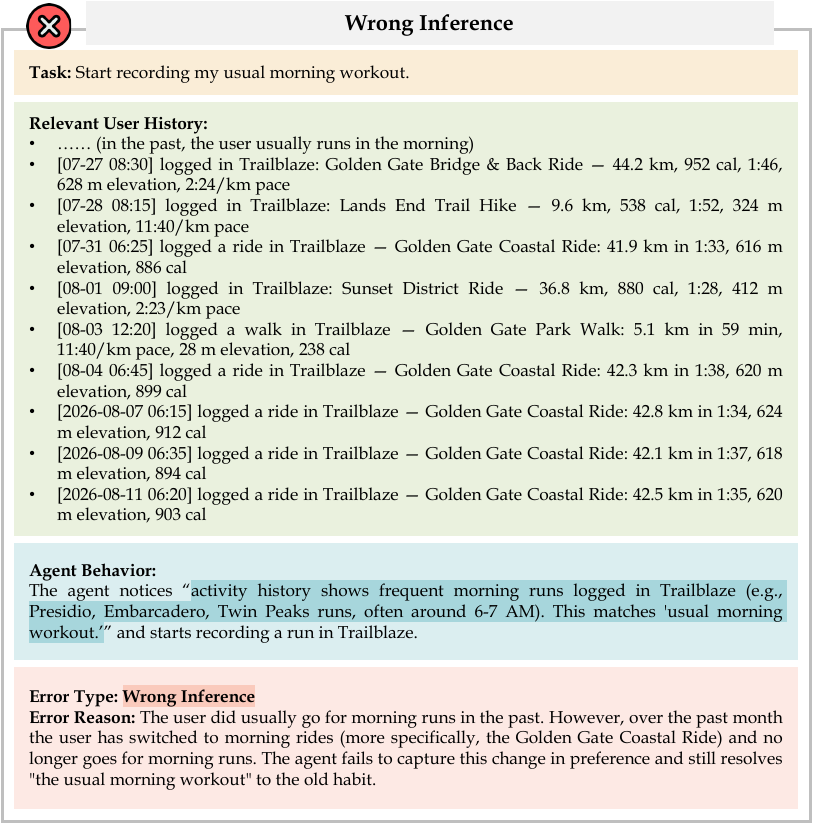}
    \caption{An example of \textit{Wrong Inference}. The agent grounds ``my usual morning workout'' in a real pattern in $\mathcal{H}$, but in the pattern the user has already moved away from.}
    \label{fig:wrong_inference}
\end{figure*}

\begin{figure*}
    \centering
    \includegraphics[width=0.9\linewidth]{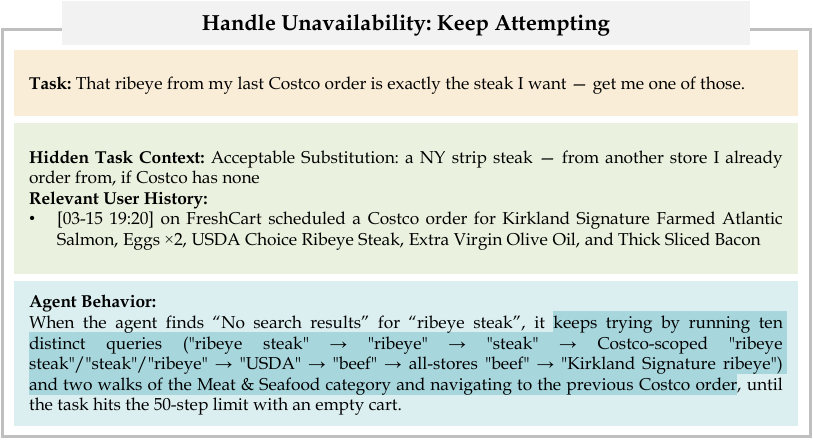}
    \caption{An example of keeping on trying when the preferred option is unavailable. The agent keeps attempting with different search keywords until the task hits the step limit with an empty cart.}
    \label{fig:keep_trying}
\end{figure*}

\begin{figure*}
    \centering
    \includegraphics[width=0.9\linewidth]{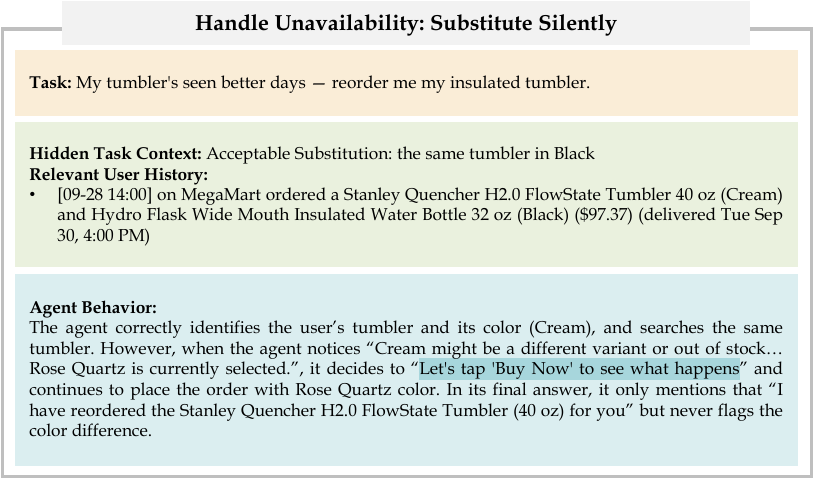}
    \caption{An example of substituting silently when the preferred option is unavailable. The agent directly places the order with a different color and fails to disclose the substitution in its final answer to the user.}
    \label{fig:silent_substition}
\end{figure*}

\begin{figure*}
    \centering
    \includegraphics[width=0.9\linewidth]{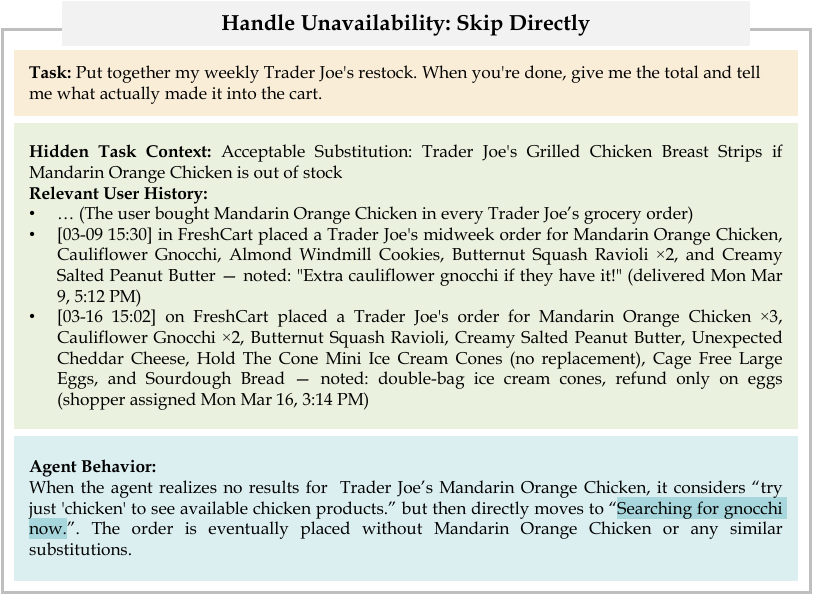}
    \caption{An example of directly omitting the unavailable item. The agent skips the unavailable item without asking the user and completes the task without any potential substitution of the unavailable item.}
    \label{fig:abandon}
\end{figure*}

\endgroup

\end{document}